\documentclass[3p,twocolumn]{elsarticle}

\usepackage{hyperref}
\usepackage{enumerate} 
\usepackage{amssymb}
\usepackage{algorithmicx,algorithm}
\usepackage{hyperref}
\usepackage{multirow}
\usepackage{stfloats}
\usepackage{graphicx}
\usepackage{color}
\usepackage[justification=centering]{caption}
\usepackage{url}
\usepackage{xcolor}
\usepackage{pdflscape}
\journal{Journal of \LaTeX\ Templates}

\begin{document}

\begin{frontmatter}

\title{IL-MCAM: An Interactive Learning and Multi-channel Attention Mechanism-based Weakly Supervised Colorectal Histopathology Image Classification Approach}

\author[1]{Haoyuan Chen}

\author[1]{Chen Li\corref{cor1}}
\cortext[cor1]{Corresponding author:} \ead{lichen201096@hotmail.com}

\author[2]{Xiaoyan Li\corref{cor1}}
\cortext[cor2]{Corresponding author:} \ead{lixiaoyan@cancerhosp-ln-cmu.com}

\author[1]{Md Mamunur Rahaman}

\author[1]{Weiming Hu}

\author[1]{Yixin Li}

\author[1]{Wanli Liu}

\author[1,3]{Changhao Sun}

\author[4]{Hongzan Sun}

\author[5]{Xinyu Huang}

\author[5]{Marcin Grzegorzek}

\address[1]{Microscopic Image and Medical Image Analysis Group, College of Medicine and Biological
Information Engineering, Northeastern University, China}

\address[2]{Department of Pathology, Cancer Hospital of China Medical University, Liaoning Cancer Hospital and Institute, China}

\address[3]{Shenyang Institute of Automation, Chinese Academy of Sciences, China}

\address[4]{Department of Radiology, Shengjing Hospital of China Medical University, China}

\address[5]{Institute of Medical Informatics, University of Luebeck, Germany}

\begin{abstract}
In recent years, colorectal cancer has become one of the most significant diseases that endanger human health. Deep learning methods are increasingly important for the classification of colorectal histopathology images. However, existing approaches focus more on end-to-end automatic classification using computers rather than human-computer interaction. In this paper, we propose an IL-MCAM framework. It is based on attention mechanisms and interactive learning. The proposed IL-MCAM framework includes two stages: automatic learning (AL) and interactivity learning (IL). In the AL stage, a multi-channel attention mechanism model containing three different attention mechanism channels and convolutional neural networks is used to extract multi-channel features for classification. In the IL stage, the proposed IL-MCAM framework continuously adds misclassified images to the training set in an interactive approach, which improves the classification ability of the MCAM model. We carried out a comparison experiment on our dataset and an extended experiment on the HE-NCT-CRC-100K dataset to verify the performance of the proposed IL-MCAM framework, achieving classification accuracies of $98.98\%$ and $99.77\%$, respectively. In addition, we conducted an ablation experiment and an interchangeability experiment to verify the ability and interchangeability of the three channels. The experimental results show that the proposed IL-MCAM framework has excellent performance in the colorectal histopathological image classification tasks.
\end{abstract}

\begin{keyword}
Colorectal cancer histopathology image \sep Attention mechanism \sep Interactivity learning \sep Image classification
\end{keyword}

\end{frontmatter}

\section{Introduction}

Cancer is a life-threatening disease caused by the over-proliferation of cells in the human body. Because cancer growth is uncontrollable and irregular, cancer can invade the surrounding tissues and rapidly metastasise to other body parts through the circulatory or lymphatic systems. Colorectal cancer (CRC) is a common type of intestinal cancer. CRC is initially a polyp, and with time it transforms into cancerous cells. CRC has a high incidence and mortality rate. Among cancers, it has the third highest incidence and second highest mortality rate in the world~\cite{ganz-2012-ASOPI,bray-2018-GCSEO}. In China, the age-standardized incidence of CRC in 2018 is 28.1 per 100,000 and 19.4 per 100,000 for men and women, respectively, and this is increasing annually~\cite{zhou-2021-CCBAT}. Therefore, it is imperative for doctors to diagnose colorectal cancer quickly and accurately.

The traditional approach for diagnosing CRC is for histopathological examination. The pathologist stains the tissue specimen with haematoxylin and eosin (H\&E), and then determines the area of malignancy by observing changes in cell morphology and tissue composition under a microscope~\cite{iftikhar-2016-ACCGP}. However, the results obtained by pathologists are often time-consuming and highly subjective, which makes histopathological evaluation by pathologists alone inadequate~\cite{aeffner-2017-TGSPI}. Therefore, the emergence of rapid and efficient computer-aided diagnosis (CAD) technology is crucial. CAD assists doctors in improving the quality and efficiency of diagnosis in medical images through image processing, pattern recognition and machine learning~\cite{ai-2021-ASRFG}.

\begin{figure*}[ht]
\centering
\includegraphics[trim={0cm 0cm 0cm 0cm},clip,width= \textwidth]{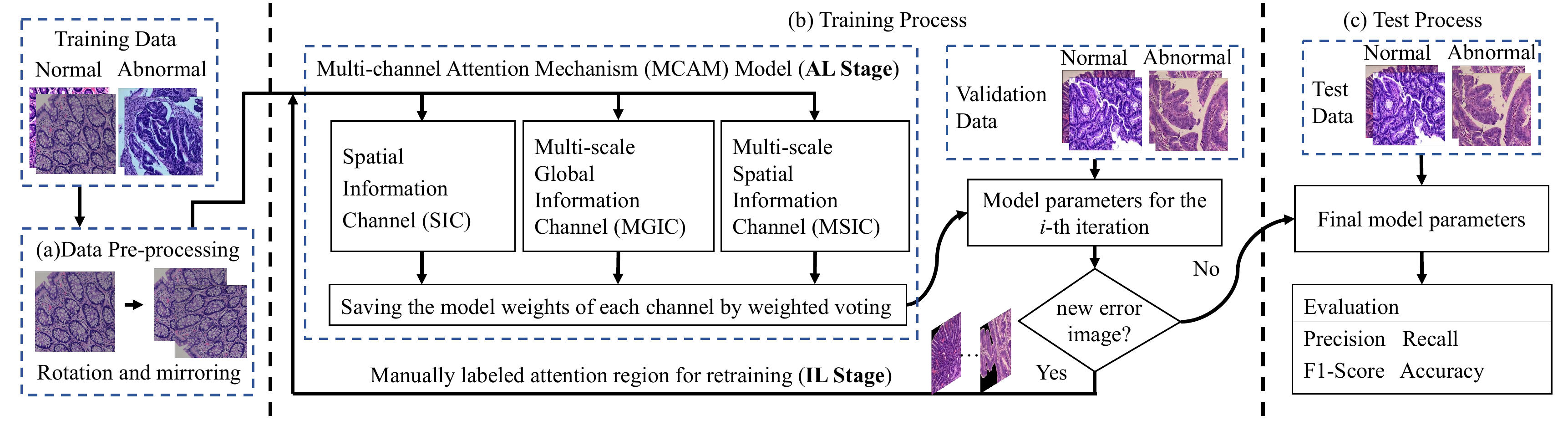}
\caption{The whole process of IL-MCAM framework.}     
\label{fig:Process-AM-IL}
\end{figure*}

Traditional CAD approaches usually use classical machine learning methods, which work as follows: First, the image features, such as shape, colour and texture, are extracted manually. Then the extracted features are classified by a classifier~\cite{jordan-2015-MLTPP}. With the advent of deep learning, the subjective extraction of features in machine learning has been replaced by automatic feature learning in computers using convolutional neural network (CNN) models, thereby dramatically increasing the accuracy and efficiency of CAD~\cite{lecun-2015-DL}. However, CNN models have a disadvantage in that they do not appropriately extract valid information from small-scale datasets. This disadvantage makes it especially important to combine CNN models with an attention mechanism (AM). AM is an approach that assign the computational resources in favour of the most informative component of the signal~\cite{itti-1998-AMOSV}. AM approaches mainly represent the automatic selection of attention regions in computer vision tasks. Different AM approaches can be separated into spatial, channel and mixed domains depending on the different priorities of computational resource allocation, leading to different AM approaches having different attention regions within the same task. In medical image datasets, complex components and limited differentiation between different stages make identification of attention regions difficult using a single AM. Consequently, we propose the IL-MCAM framework: a weakly supervised learning approach based on multi-channel attention mechanism (MCAM) and interactive learning (IL) to improve accuracy in colorectal histopathological image classification (CHIC) tasks. The whole process of this approach is shown in Fig.~\ref{fig:Process-AM-IL}.

The training process is separated into two stages: automatic learning (AL) stage and IL stage. The AL stage is performed with the MCAM model, consisting of three channels: spatial information channel (SIC), multi-scale global information channel (MGIC) and multi-scale spatial information channel (MSIC). The training images are input into the MCAM model to obtain the model parameters of the AL stage using the weighted voting approach after several epochs. In the IL stage, the validation images are input into the previously trained model for classification. Misclassified images are manually labelled with attention regions and then  added to the training set for retraining. This process is iterated several times until no new errors are generated in the IL stage. Finally, the model parameters of the final iteration are preserved, and the test images are input to obtain the CHIC task results.

The contributions of this paper are as follows:
\begin{itemize}
\item First, the MCAM model can identify the attention regions as accurately as possible in the channel and spatial dimensions by integrating different attention mechanisms, thereby improving the accuracy of the CHIC task in the AL stage.
\item Second, the IL approach manually labels attention regions, enabling modification of the errors caused by the MCAM model in the AL stage, which further improves the accuracy of the CHIC task in the IL stage.
\end{itemize}

This paper is organised as follow: Section~\ref{section:rw} provides a review of the status of CHIC tasks in the past few years, Section~\ref{section:mt} details the approaches in this paper, Section~\ref{section:ea} presents experimental results, Section~\ref{section:ds} analyses the reasons based on the experimental results. Finally, Section~\ref{section:c} concludes the paper with a brief conclusion.

\section{Related Work}
\label{section:rw}
\subsection{Classification tasks in Colorectal Histopathology Research}
In CHIC tasks, there are several examples of machine learning approaches for classification by manually extracting image features into classifiers. In \cite{linder-2012-IOTEA}, local binary pattern (LBP) texture features with an integrated contrast measured through a support vector machine (SVM) classifier obtained a $99.5\%$ accuracy in normal-abnormal binary classification of 643 patient-level images. In \cite{jiao-2013-CCDUW}, 60 normal and abnormal images are extracted with mean and variance features and grey-level co-occurrence matrix (GLCM) features. The extraction features are used for classification by the SVM classifier, and an $89.5\%$ F1-score in binary classification is obtained after 3-fold cross-validation. In \cite{peyret-2015-TAFCT}, the experimental process is as follows: The texture features, including LBP, Haraclick features and local intensity order patterns, are dimensionally reduced dimensional using principal component analysis, and then different classifiers are used to classify the reduced dimensional features. The experiment achieves $91.3\%$ accuracy for 464 images using the SVM classifier. In \cite{chaddad-2016-MTAOC}, the integration features, including three different texture features, Laplacian-of-Gaussian filter, discrete wavelets and GLCM, are used for classification in a linear discriminant analysis classifier. An accuracy of $98.2\%$ is obtained for 480 colorectal histopathology images. In \cite{kather-2016-MCTAI}, the histogram-low features of 5000 patch-level images using the rbf-SVM classifier obtained $98.6\%$ accuracy in the binary classification problem.

In recent years, with the eminence of deep learning in regular image classification tasks, an increasing amount of research has been conducted on CHIC tasks. In \cite{sari-2018-UFEVD}, a semi-supervised classification approach based on restricted Boltzmann machines (RBMs) is proposed. This approach uses the features of the sub-regions in the image for learning. A deep belief network of consecutive RBMs is constructed to extract pixels. The activation values of the hidden unit nodes of the RBM are used as the final features. The extracted features are learned using an unsupervised clustering approach. Two datasets containing 3236 and 1644 images is used in multi-classification and obtained accuracies of $96.11\%$ and $78.99\%$, respectively. In \cite{malik-2019-CCDFH}, a new adaptive CNN implementation model that performs well even in low-resolution and constrained images is proposed. Using this approach, an accuracy of $94.5\%$ is achieved for 3200 patch-level images. In \cite{yoon-2019-TIICH}, a CNN based on a modified VGG model is proposed for CRC classification. An accuracy of $82\%$ is achieved for 10280 images. In \cite{dif-2020-ANDLM}, a dynamic ensemble learning method is proposed for a multiclass CHIC task. This approach first uses transfer learning to train each model and then a particle swarm optimisation algorithm to select and integrate the models. It obtains an accuracy of $94.52\%$ for 5000 patch-level histopathology images using the ResNet-121 architecture. In \cite{nguyen-2020-AEDLA}, a multi-classification accuracy of $95.3\%$ is achieved for 410 patient-level images using a combination of classical CNN and CapsNet models. In \cite{ghosh2021colorectal}, an ensemble model based on Xception, DenseNet-121 and InceptionResNet-V2 achieves $92.83\%$, $96.16\%$ and $99.13\%$ accuracy for CRC-5000, NCT-CRC-HE-100K and the merged datasets, respectively. Similarly, in \cite{hamida2021deep}, the ResNet model using fine-tuning achieves $96.77\%$, $99.76\%$ and $99.98\%$ accuracy for CRC-5000, NCT-CRC-HE-100K and the merged datasets, respectively. In \cite{ohata2021novel}, a $92.083\%$ accuracy is obtained by evaluating 108 different combinations of features and classifiers on the CRC-5000 dataset. In \cite{sarkar2021classification}, an encoder unit of an autoencoder module and a modified DenseNet-121 architecture are used for the purpose approach. This approach, has an accuracy of $97.2\%$ for the Zenodo-100K colorectal histopathology dataset. In \cite{sarwinda2021deep}, the ResNet-50 model is used on private datasets and obtains an overall accuracy higher than $80\%$.
\subsection{Overview of Deep Learning Methods}
In computer vision tasks, CNN models are the most used deep learning methods. The continuous improvement of transformer models and multilayer perceptron (MLP) models has made them popular. Especially, deep learning methods are widely used in many biomedical image analysis tasks, such as COVID-19 idetificaion~\cite{rahaman2020identification,li2020sars}, microorganism image analysis~\cite{kosov2018environmental,li2019survey,zhang2021lcu}, histopatholgical image analysis~\cite{zhou2020comprehensive,xue2020application,chen2021gashis,li2022hierarchical,hu2022gashissdb} , cytopathological image analysis~\cite{rahaman2020survey,mamunur2021deepcervix,liu2021aspect} and sperm image analysis~\cite{li2020foldover,chen2022svia}.

The first application of the CNN model was LeNet, proposed by LeCun et al. in 1989~\cite{lecun-1989-BATHZ}. In 2012, Krizhevsky et al. proposed AlexNet, which that uses the powerful parallel computing ability of the graphics processing unit to process several matrix operations during training~\cite{krizhevsky-2012-ICWDC}. Since then, deep learning methods have formally replaced traditional machine learning methods. The subsequent improvements for the CNN models focus on three aspects: network depth, network width and hybrid network depth and width. The VGG~\cite{simonyan-2014-VGG}, ResNet~\cite{he-2016-DRLFI} and DenseNet~\cite{huang-2017-densenet} models increase the network depth by using small convolutional layers, residual mechanisms and dense layers to improve model performance.  The Inception-V3~\cite{szegedy-2016-inception} and Xception~\cite{chollet-2017-xception} models enhance the network width by using multi-scale inception blocks and separable convolutional blocks. Some models such as InceptionResNet~\cite{szegedy-2017-IIATI} and ResNeXt~\cite{xie-2017-ARTFD} enhance the network's depth and width by combining inception blocks and residual mechanisms in the feature extraction layer of the network, thereby improving the classification task performance.

Transformer models were initially proposed by Vaswani et al. in 2017 for natural language processing tasks~\cite{vaswani-2017-AIAYN}. In recent years they have moved to computer vision tasks. Transformer models are divided into two main categories, pure transformer models and transformer models combined with CNN~\cite{khan-2021-TIVAS}. Pure transformer models include ViT~\cite{dosovitskiy-2020-AIIWW}, DeiT~\cite{touvron-2021-TDITD}, CaiT~\cite{touvron-2021-GDWIT} and T2T-ViT~\cite{yuan-2021-T2TVIT} models. These models directly input the image into the transformer encoder after position encoding. Transformer models combining with CNNs are the BoTNet~\cite{srinivas-2021-BTFVR}, CoaT~\cite{xu-2021-CCIT} and LeViT~\cite{graham-2021-levit} models, which input the feature maps obtained by convolution of the images into the transformer encoder.

The currently implemented MLP models are the improved versions of transformer models. The MLP-mixer~\cite{tolstikhin-2021-MAAAF} model is improved by replacing the self-attention layers of the ViT model with several perceptrons. The gMLP~\cite{liu-2021-PATM} and ResMLP~\cite{touvron-2021-RFNFI} models add a gate mechanism and a residual mechanism to the MLP-mixer model to improve the performance.

\section{IL-MCAM framework}
\label{section:mt}
\begin{figure*}[hbp]
\centering
\includegraphics[trim={0cm 0cm 0cm 0cm},clip,width= \textwidth]{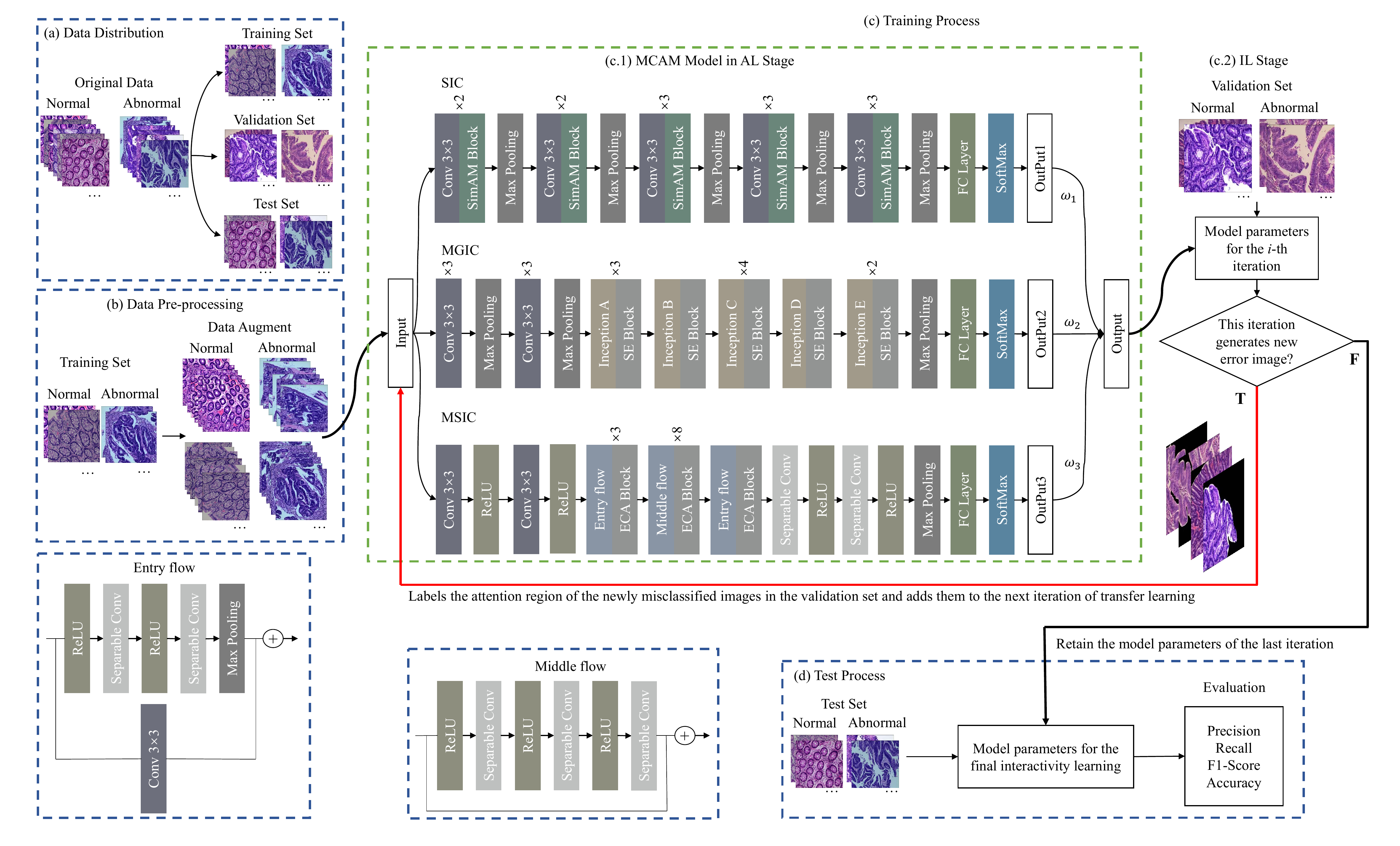}
\caption{The structure of the proposed IL-MCAM framework.}
\label{fig:Structure-AM-IL}
\end{figure*}
The whole process of the IL-MCAM framework is shown in Fig.~\ref{fig:Structure-AM-IL}.

Step 1: In Fig.~\ref{fig:Structure-AM-IL}-(a), the original images are proportionally divided into training, validation and test sets.

Step 2: In Fig.~\ref{fig:Structure-AM-IL}-(b), the training set images are data augmented using the rotation and expansion approach.

Step 3: In Fig.~\ref{fig:Structure-AM-IL}-(c.1), the AL stage is implemented by inputting the training set images into the MCAM model for training. The MCAM comprises of three parallel channels, and each channel is composed of CNN models with integrated attention mechanisms, namely SIC, MGIC, and MSIC. The differences between the three different attention mechanism channels and their corresponding CNN models are listed in Table \ref{Table:difference}. SIC consists of the VGG-16~\cite{simonyan-2014-VGG} model-integrated mixed-domain SimAM~\cite{yang-2021-simam} to extract spatial image information, and each SimAM block is added to the VGG-16 model after each convolutional layer. MGIC consists of the Inception-V3~\cite{szegedy-2016-RTIAF} model integrated channel domain SE~\cite{hu-2018-SE} to extract image multi-scale global channel information, and each SE block is added to the Inception-V3 model after each inception block. MSIC consists of the Xception~\cite{chollet-2017-XDLWD} model integrated low-cost channel domain ECA~\cite{wang-2020-eca} to extract image multi-scale local channel information, and each ECA block is added to the Xception model after each flow block. The three trained channels are distributed using the weighted voting approach for each channel component weight.

Step 4: In Fig.~\ref{fig:Structure-AM-IL}-(c.2), the IL stage uses a human-machine interaction. The misclassified images in the validation set are labelled with attention regions and then input into the training set for retraining with a transfer learning approach. This process is repeated for several iterations until there are no new misclassified images in the validation set. The model parameters of the last iteration are reserved as the final output.

Step 5: In Fig.~\ref{fig:Structure-AM-IL}-(d), the test set images are tested using the last reserved model parameters to obtain the final classification results.

This section is composed as follows: CNNs are explained in Section~\ref{subsection:CNN}, the transfer learning approach is described in Section~\ref{subsection:TL}, building a MCAM model is detailed in Section~\ref{subsection:MCAM}, and interactive learning strategy is explained in Section~\ref{subsection:IL}.

\begin{table*}[ht]
\scriptsize
\centering
\caption{The differences between three different attention mechanism channels and their corresponding CNN models. The bottom row shows classification layers and other rows are feature extraction layers.}
\label{Table:difference}
\begin{tabular}{cc|cc|cc}
\hline
VGG-16               & SIC                   & Inception-V3         & MGIC                  & Xception             & MSIC                 \\ \hline
Conv                 & Conv                  & Conv                 & Conv                  & Conv                 & Conv                 \\
                     & SimAM Block           &                      &                       &                      &                      \\ \hline
Conv                 & Conv                  & Conv                 & Conv                  & ReLU                 & ReLU                 \\
                     & SimAM Block           &                      &                       &                      &                      \\ \hline
Maxpool              & Maxpool               & Maxpool              & Maxpool               & Conv                 & Conv                 \\
\multicolumn{1}{l}{} & \multicolumn{1}{l|}{} & \multicolumn{1}{l}{} & \multicolumn{1}{l|}{} &                      &                      \\ \hline
Conv                 & Conv                  & Conv                 & Conv                  & ReLU                 & ReLU                 \\
                     & SimAM Block           &                      &                       &                      &                      \\ \hline
Conv                 & Conv                  & Conv                 & Conv                  & Entry flow           & Entry flow           \\
                     & SimAM Block           &                      &                       &                      & ECA Block            \\ \hline
Maxpool              & Maxpool               & Maxpool              & Maxpool               & Entry flow           & Entry flow           \\
\multicolumn{1}{l}{} & \multicolumn{1}{l|}{} & \multicolumn{1}{l}{} & \multicolumn{1}{l|}{} &                      & ECA Block            \\ \hline
Conv                 & Conv                  & Inception A          & Incecption A          & Entry flow           & Entry flow           \\
                     & SimAM Block           &                      & SE Block              &                      & ECA Block            \\ \hline
Conv                 & Conv                  & Inception A          & Inception A           & Middle flow          & Middle flow          \\
                     & SimAM Block           &                      & SE Block              & \multicolumn{1}{l}{} & ECA Block            \\ \hline
Conv                 & Conv                  & Inception A          & Inception A           & Middle flow          & Middle flow          \\
                     & SimAM Block           &                      & SE Block              & \multicolumn{1}{l}{} & ECA Block            \\ \hline
Maxpool              & Maxpool               & Inception B          & Inception B           & Middle flow          & Middle flow          \\
\multicolumn{1}{l}{} & \multicolumn{1}{l|}{} & \multicolumn{1}{l}{} & SE Block              & \multicolumn{1}{l}{} & ECA Block            \\ \hline
Conv                 & Conv                  & Inception C          & Inception C           & Middle flow          & Middle flow          \\
                     & SimAM Block           & \multicolumn{1}{l}{} & SE Block              & \multicolumn{1}{l}{} & ECA Block            \\ \hline
Conv                 & Conv                  & Inception C          & Inception C           & Middle flow          & Middle flow          \\
                     & SimAM Block           & \multicolumn{1}{l}{} & SE Block              & \multicolumn{1}{l}{} & ECA Block            \\ \hline
Conv                 & Conv                  & Inception C          & Inception C           & Middle flow          & Middle flow          \\
                     & SimAM Block           & \multicolumn{1}{l}{} & SE Block              & \multicolumn{1}{l}{} & ECA Block            \\ \hline
Maxpool              & Maxpool               & Inception C          & Inception C           & Middle flow          & Middle flow          \\
\multicolumn{1}{l}{} & \multicolumn{1}{l|}{} & \multicolumn{1}{l}{} & SE Block              & \multicolumn{1}{l}{} & ECA Block            \\ \hline
Conv                 & Conv                  & Inception D          & Inception D           & Middle flow          & Middle flow          \\
                     & SimAM Block           & \multicolumn{1}{l}{} & SE Block              & \multicolumn{1}{l}{} & ECA Block            \\ \hline
Conv                 & Conv                  & Inception E          & Inception E           & Entry flow           & Entry flow           \\
                     & SimAM Block           & \multicolumn{1}{l}{} & SE Block              & \multicolumn{1}{l}{} & ECA Block            \\ \hline
Conv                 & Conv                  & Inception E          & Inception E           & Separable Conv       & Separable Conv       \\
                     & SimAM Block           & \multicolumn{1}{l}{} & SE Block              & \multicolumn{1}{l}{} & \multicolumn{1}{l}{} \\ \cline{5-6} 
\multicolumn{1}{l}{} & \multicolumn{1}{l|}{} & \multicolumn{1}{l}{} & \multicolumn{1}{l|}{} & ReLU                 & ReLU                 \\
\multicolumn{1}{l}{} & \multicolumn{1}{l|}{} & \multicolumn{1}{l}{} & \multicolumn{1}{l|}{} & \multicolumn{1}{l}{} & \multicolumn{1}{l}{} \\ \cline{5-6} 
\multicolumn{1}{l}{} & \multicolumn{1}{l|}{} & \multicolumn{1}{l}{} & \multicolumn{1}{l|}{} & Separable Conv       & Separable Conv       \\
\multicolumn{1}{l}{} & \multicolumn{1}{l|}{} & \multicolumn{1}{l}{} & \multicolumn{1}{l|}{} & \multicolumn{1}{l}{} & \multicolumn{1}{l}{} \\ \cline{5-6} 
\multicolumn{1}{l}{} & \multicolumn{1}{l|}{} & \multicolumn{1}{l}{} & \multicolumn{1}{l|}{} & ReLU                 & ReLU                 \\
\multicolumn{1}{l}{} & \multicolumn{1}{l|}{} & \multicolumn{1}{l}{} & \multicolumn{1}{l|}{} & \multicolumn{1}{l}{} & \multicolumn{1}{l}{} \\ \hline
Maxpool              & Maxpool               & Maxpool              & Maxpool               & Maxpool              & Maxpool              \\
FC Layer             & FC Layer              & FC Layer             & FC Layer              & FC Layer             & FC Layer             \\
Softmax              & Softmax               & Softmax              & Softmax               & Softmax              & Softmax              \\ \hline
\end{tabular}
\end{table*}
%\end{landscape}
\subsection{Convolutional Neural Network (CNN)}
\label{subsection:CNN}
A CNN is a feedforward neural network that includes the computation of the convolution and depth structure. A CNN consists of several layers, including a convolution layer, pooling layer, and fully connected layer. The convolutional layer, which is the core of the CNN extracts image features using a convolution kernel. The pooling layer is used to compress the input feature map and extract the main features. The fully connected layer connects all features and classifies the output features using a classifier. In a CNN, the information extracted by the convolution layers of different networks is separated into two main categories: global and local. Global information refers to the macroscopic representation of an image in its class and is usually extracted by large convolution kernels and positional coding. Local information, also described as spatial information, represents the features of a restricted region of the image in its class and is usually extracted by a small convolution kernel.

The Visual Geometry Group (VGG) proposed the VGG-16~\cite{simonyan-2014-VGG} model at the University of Oxford in 2014. Its novelty contribution is raising the depths of networks from eight to 16, and converting large convolution kernels such as $7\times7$ and $5\times5$ into two or three $3\times3$ small convolution kernels. It is another milestones in deep learning after AlexNet~\cite{krizhevsky-2012-ICWDC} and the baseline for comparing new methods in the field of deep learning. The VGG model has significant advantages. It uses the small convolution kernel to enhance the extraction of spatial information better~\cite{rawat-2017-DCNNF}.

The Inception-V3~\cite{szegedy-2016-RTIAF} model is another method that modifies AlexNet~\cite{krizhevsky-2012-ICWDC} and is based on GoogLeNet~\cite{szegedy-2015-GDWC}, proposed by Szegedy et al. in 2015. Instead of using the conventional method to increase the number of network layers, the Inception-V3 model uses a novel convolution method to decompose large filter sizes by using parallel convolution and factorised convolution. The entire decomposition module is called the inception structure. Moreover, this model has five different inception structures, each with its own set of components. The Inception-V3 model uses an inception module instead of a large convolution kernel and a global average pooling layer instead of a fully connected layer, to substantially reduce the number of parameters compared with other models. Among CNN models, Inception-V3 has an exceptional ability to extract global multiscale information owing to its parallel convolution structure and partially large convolution kernels.

The Xception~\cite{chollet-2017-XDLWD} model improves the Inception-V3~\cite{szegedy-2016-RTIAF} model by combining the depth-separable convolution~\cite{sifre-2014-RSFTC} and residual mechanism~\cite{he-2016-DRLFI}. Unlike the standard convolution approach, depth-separable convolution is performed separately for each channel in the feature map~\cite{sifre-2014-RSFTC}. The advantage of Xception is its combination of the residual structure and depth-separable convolution. The depth-separable convolution effectively extracts the multi-scale features of the image, and the residual mechanism makes the network model converge easily. In contrast to the Inception-V3 model, the small convolutional kernel in depth-separable convolution gives the Xception model a good local multi-scale information extraction ability.
\subsection{Transfer Learning (TL)}
\label{subsection:TL}
Training CNN models from scratch requires a large amount of data and high computational power, resulting in a longer training time. Moreover, the small size and vague labels of medical datasets make TL critical for CHIC tasks~\cite{raghu-2019-UTLFM}. TL is a machine learning approach wherein a pre-trained model is reused in another task~\cite{pan-2009-ASOTL}. The TL process has two steps: the first is selecting an original dataset and pre-training on the original dataset. The second is to fine-tune the pre-trained model using the dataset of the target task.

In this paper, we use the ImageNet dataset as the original dataset to pre-train the model. Because low workstation computing power makes it challenging to pre-train MCAM models directly using the TL approach, we make changes based on traditional transfer learning. The pre-training parameters of the traditional VGG-16, Inception-V3, and Xception models in the Pytorchvision package are loaded layer by layer according to the same parts of SIC, MGIC and MSIC as in Table \ref{Table:difference}, which are frozen in training. Only AM layers and fully connected layers are used for fine-tuning, and the weights of each channel are assigned using weighted voting.
\subsection{Multi-Channel Attention Mechanisms}
\label{subsection:MCAM}
AMs are inspired by human biological systems. They tend to focus on specific parts when processing large sets of information~\cite{itti-1998-AMOSV}. The AM approach has become one of the most imperative concepts in the field of deep learning~\cite{niu-2021-AROAM}. However, traditional AMs have several disadvantages. For example, a single AM can identify redundant information, leading to mistakes. To overcome some reduce the disadvantages of AMs, we propose an MCAM model. This model extracts features from multiple perspectives through three channels: SIC, MGIC and MSIC. These three complementary channels improve the precision of identifying attention regions and the accuracy of classification tasks.

\begin{figure}[ht]
\centering
\includegraphics[trim={0cm 0cm 0cm 0cm},clip,width= 0.5\textwidth]{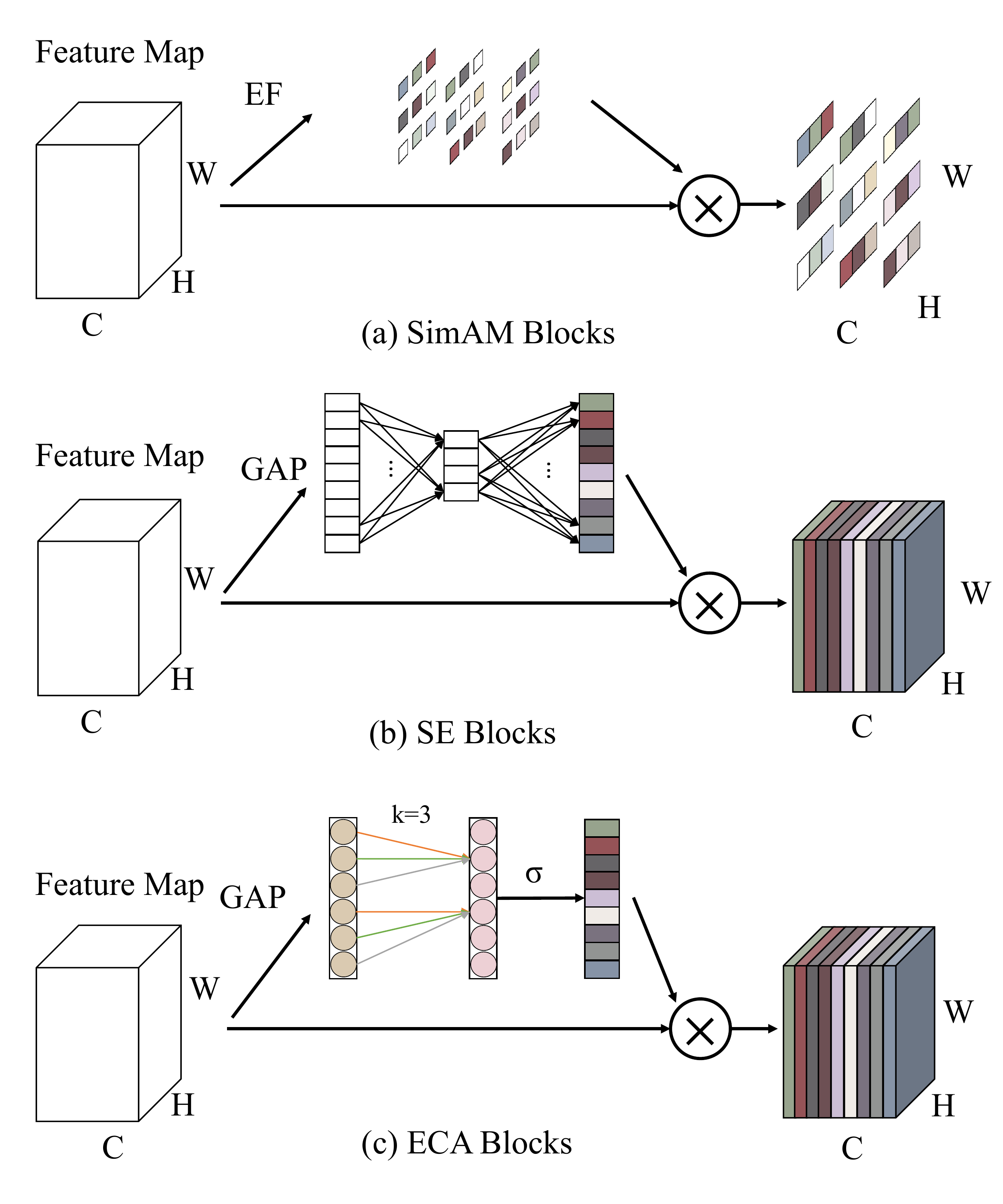}
\caption{The structure of three different attention mechanism. (a) is SimAM Blocks after each convolution layer in SIC. (b) is SE Blocks after each Inception block in MGIC. (c) is ECA blocks after each flow in MSIC.}
\label{fig:Structure-Attention}
\end{figure}

\emph{SIC:} SIC is expectation because its capability of extracting spatial information is excellent. The SimAM attention mechanism has an outstanding ability to distribute weights to the features of spatial dimensions~\cite{yang-2021-simam}. The structure of the SimAM attention mechanism is illustrated in Fig.~\ref{fig:Structure-Attention}-(a). In visual neuroscience, the most informative neurones show different firing patterns in the surrounding neurons and keep the activity of the surrounding neurons, a phenomenon known as spatial suppression~\cite{webb2005early}. The easiest way to find these spatially suppressed neurones is to measure the linear separability between the target and other neurones. In computer vision tasks, the edge features of images often play a crucial role in classification tasks. Furthermore, the edge features of images are the same as those of spatial suppression neurones, which often exhibit incredibly high contrast with the surrounding colours and textures. Therefore, the SimAM attention mechanism works by using an energy function (EF) from neuroscience to assign weights to different spatial locations. The energy function perceives each pixel of the feature map as a neuron and the minimal energy of neurones can be expressed as following:
\begin{equation}
  e^*_t=\frac{4(\sigma ^2+\lambda )}{(t-\mu )^2+2\sigma ^2+2\lambda }\\
  \label{eq:1}
\end{equation}
where $t$ is target neuron, $i$ is index over spatial dimension, $\mu =\frac{1}{M}\sum_{i=1}^{M}x_i$ and $\sigma ^2=\frac{1}{M}\sum_{i=1}^{M}(x_i-\mu )^2$ are mean and variance calculated over all neurons except $t$ in that channel, $x_i$ is other neurons in a same channel, $M=H \times W$ is number of neurons on the channel and $\lambda$ is a coefficient and is set to $1e-4$ according to the experiments on the CIFAR datasets~\cite{yang-2021-simam}. Spatially suppression neurons are less similar to other neurons and exhibit a high linear separability, thus showing a significant deviation in $t$ and $u$ leading to a low $e^*_t$. Meanwhile, in neuroscience, it is considered that lower energy indicates neurons that are more differentiated from surrounding neurons. Therefore, the weights of each neuron can be calculated from $e^*_t$. The optimization phase of the whole SimAM attention mechanism is obtained by scaling operator:
\begin{equation}
  \widetilde{X}=sigmoid(\frac{1}{E})\cdot X\\
  \label{eq:2}
\end{equation}
where $X$ and $\widetilde{X}$ are input feature map and output feature map, $E$ is all $e^*_t$ are grouped in spatial and channel dimensions. Finally, the confidence of each neuron at each place is obtained by sigmoid activation function.

In Section~\ref{subsection:CNN}, it is shown that the VGG-16 model is capable of extracting spatial information~\cite{simonyan-2014-VGG}. Above all, the SimAM attention mechanism after every convolutional layer in VGG-16 model is designed to extract spatial information in the SIC.

\emph{MGIC:} In the MGIC, the model is expected to be capable of extracting multi-scale global information. In Section~\ref{subsection:CNN}, it explained that --- the Inception-V3 model is the best CNN model that can extract global information~\cite{szegedy-2016-RTIAF}. Therefore, the Inception-V3 model is selected to extract features in MGIC. The extraction of multi-scale information in the Inception-V3 model is implemented by concatenating different sized receptive fields, so the multi-scale ability of the Inception-V3 model is represented in the channel domain of each feature map. The SE attention mechanism, which possesses a good distribution of channel weights, is selected to strengthen the importance between the channel features in the MGIC~\cite{hu-2018-SE}. The structure of the SE attention mechanism is shown in Fig.~\ref{fig:Structure-Attention}-(b). The SE attention mechanism consists of two phases: squeeze phase and excitation phase. The SE attention mechanism consists of two phases: squeeze and excitation. The squeeze phase encodes the entire spatial features into a global feature by global average pooling to generate channel-wise statistics. The excitation phase obtains the channel-wise importance using two fully connected layers, a dimensionality-reduction layer and a dimensionality-increasing layer, and the final channel-wise weights are obtained by the sigmoid activation function.

\emph{MSIC:} This channel is implemented using depth-separable convolution of the Xception~\cite{chollet-2017-XDLWD} model. Depth-separable convolution causes information extracted from each channel of the feature map to be diversified so that multi-scale spatial information can be extracted appropriately. The ECA attention mechanism is used after each flow of the Xception model to strengthen its ability to extract multi-scale information. The ECA attention mechanism uses a low time consumption to assign weights to the importance of the channel information of each feature map~\cite{wang-2020-eca}. The structure of the ECA attention mechanism is illustrated in Fig.~\ref{fig:Structure-Attention}-(c). The ECA attention mechanism first uses global average pooling (GAP) to obtain channel-wise information, then uses $1D$ convolutional that captures cross-channel interaction information with a convolutional kernel of size $k$, and finally obtains channel-wide weight information using a sigmoid activation function.

\emph{Multi-channel fusion approach:} This approach is an integrated classifier that relies on the classification decision values of different channels and the weights of each channel to improve the classification performance~\cite{mamunur2021deepcervix}. In this experiment, the last feature maps of SIC, MGIC and MSIC are used to obtain the classification decision values for each channel using pooling, fully connected and softmax layers. Then, the classification decision values of each channel are weighted and evaluated using grid weighted voting to obtain the classification decision values of the MCAM model. Finally, the category that belongs to the maximum classification decision values of the MCAM model is used as the final classification result. The calculation formula is described as follows: 
\begin{equation}
  C = \max \limits_{1<x<n}\sum_{i=1}^{3}\omega _iD_i\\
  \label{eq:3}
\end{equation}
where $C$ is classification category, $n$ is the number of categories, $i$ is the number of channels of the MCAM model and $D_i=\left \{ d_1,d_2,...,d_n \right \}$ is $n$ classification decision values of $i$th channel of MCAM model.

\subsection{Interactive Learning (IL)}
\label{subsection:IL}
\begin{figure}[ht]
\centering
\includegraphics[trim={0cm 0cm 0cm 0cm},clip,width= 0.5\textwidth]{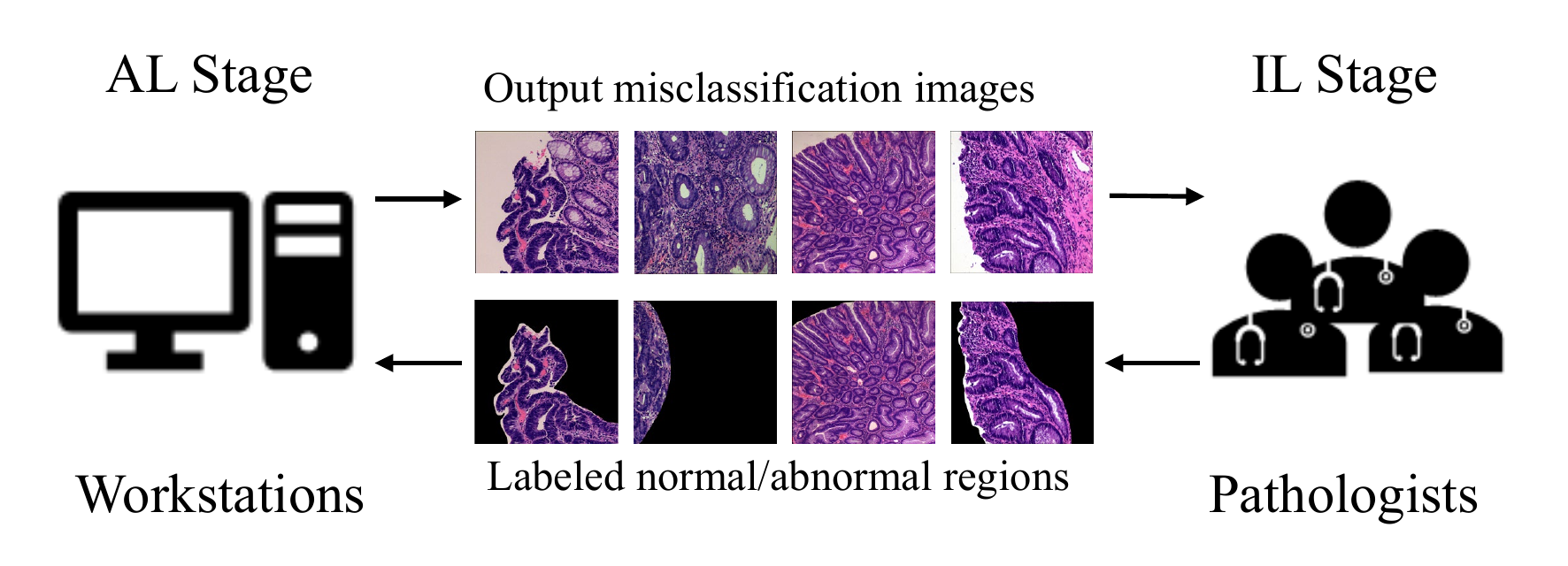}
\caption{The process of IL-MCAM framework.}
\label{fig:Process-IL}
\end{figure}

The substance of the proposed IL-MACM framework is limited frequency incremental learning. Incremental learning means that a learning system can continuously learn from new samples and can preserve most of previously acquired knowledge~\cite{castro-2018-ETEIL}. The implementation of incremental learning is achieved through an IL strategy. The IL-MACM framework process is shown in Fig.~\ref{fig:Process-IL}. First, the misclassified images in the validation set are sent to the pathologist after one training iteration. The attention regions are then discreetly and meticulously labelled by pathologists. Finally, the labelled images are input into the training set for the next training iteration until no new errors appear in the validation set.

\section{Experiment Results and Analysis}\label{section:ea}
\subsection{Experimental Settings}
\subsubsection{Dataset}
\begin{figure}[ht]
\centering
\includegraphics[trim={0cm 0cm 0cm 0cm},clip,width= 0.48\textwidth]{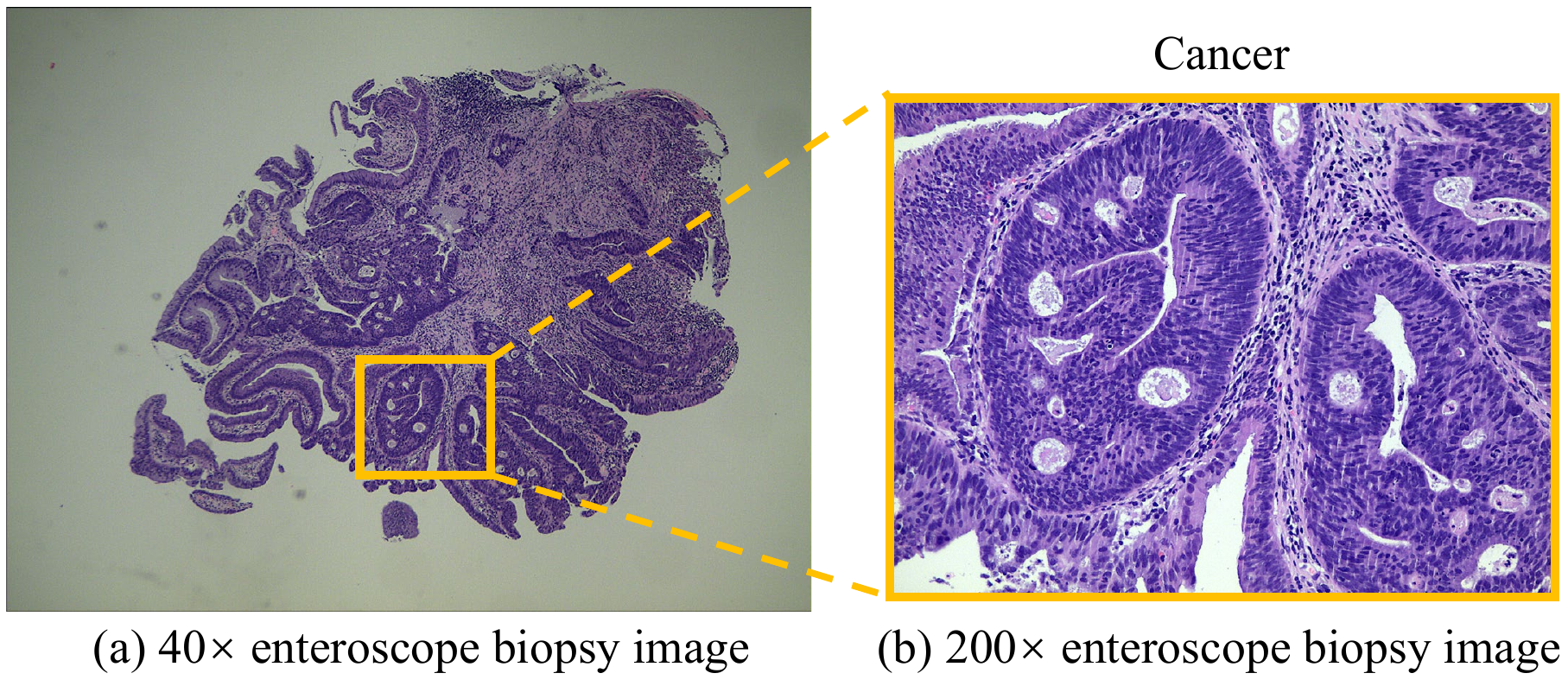}
\caption{Details of the acquisition of HE-CRC-DS. (a) is a 40$\times$ image obtained by enteroscopy biopsy. (b) is a 200$\times$ image containing in HE-CRC-DS. Pathologists first assess the most severely differentiation stage in the 40$\times$ images. Then the dataset images are obtained by adjusting the magnification to 200$\times$ and giving image-level labels according to the most severe differentiation stage.}
\label{fig:Example-enteroscope}
\end{figure}

In this study, an haematoxylin and eosin (H\&E) stained colorectal cancer (CRC) histopathology dataset (HE-CRC-DS) is used in the experiment to evaluate the classification performance of the proposed IL-MCAM approach. This dataset is collected and labelled by two pathologists from the Cancer Hospital of China Medical University and four biomedical researchers from Northeastern University. Pathologists provide electron microscopy images of histopathological sections of CRC enteroscopic biopsies using an ``Olympus'' microscope and the ``NewUsbCamera'' software and also provide image-level annotations of weakly supervised leaning processes. Biomedical researchers organise and create datasets. Details of the acquisition of HE-CRC-DS are shown in Fig.~\ref{fig:Example-enteroscope}, and the image-level labels are given as follows: First, when the pathologist finds only the differentiation stage in a 40$\times$ image, it is magnified to 200$\times$ for preservation, and this differentiation stage is then used as the image-level label. Then, if the physician finds multiple differentiation stages or similar differentiation stages in a 40$\times$ image, the most severe stage is magnified to 200$\times$ and saved, and the most severe stage is used as the image level label. In summary, the image-level label is the same as the patient-level label in HE-CRC-DS.

\begin{figure*}[ht]
\centering
\includegraphics[trim={0cm 0cm 0cm 0cm},clip,width= 0.8\textwidth]{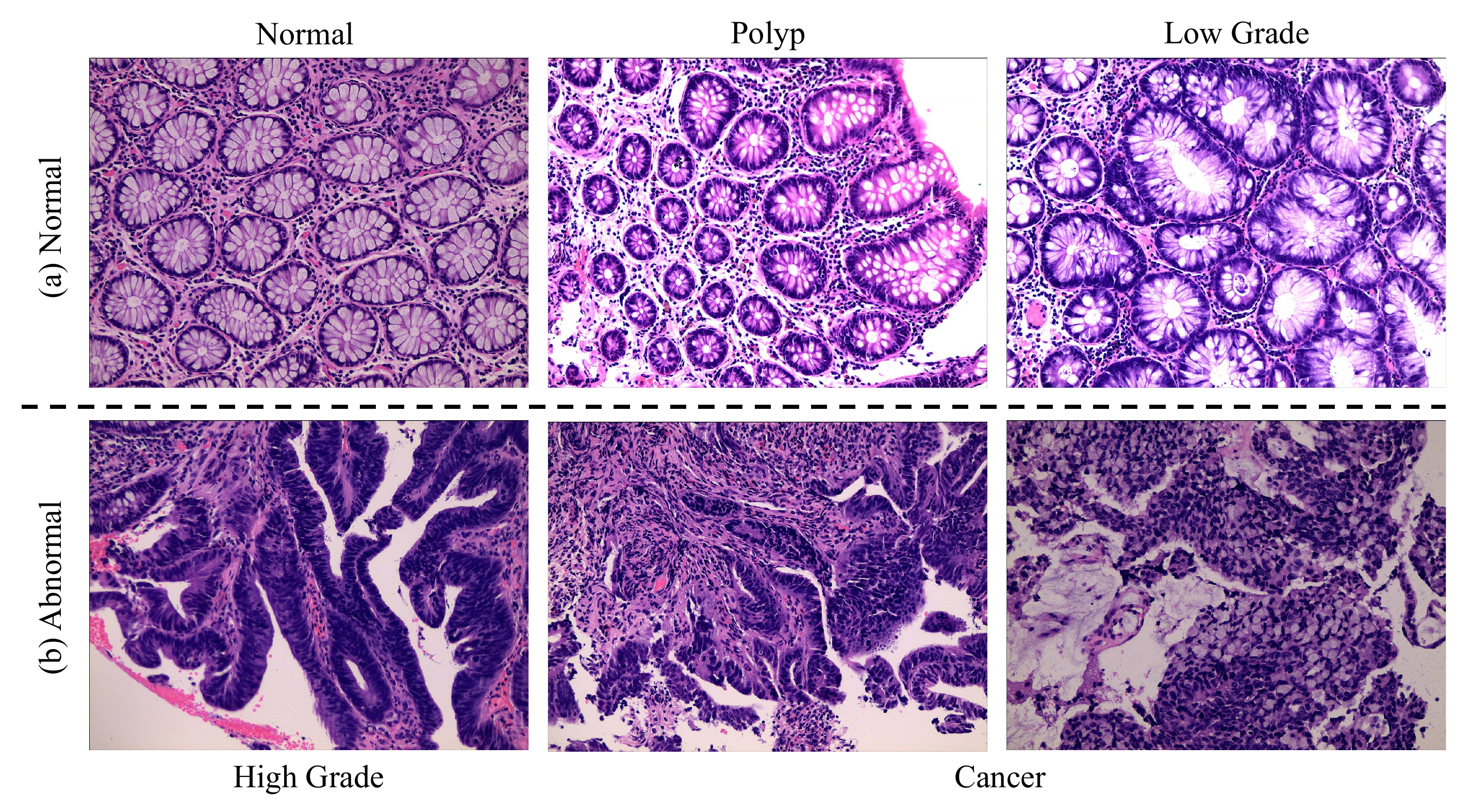}
\caption{Some examples in the HE-CRC-DS.}
\label{fig:Example-CRC}
\end{figure*}

HE-CRC-DS includes 4005 images of 2048$\times$1536 pixels in the ``.png'' file format. The overall magnification of all images in the HE-CRC-DS is 200$\times$. Most pathologists classify CRC into five categories: normal, polyp, low grade, high grade, and cancer. Due to the unbalanced data distribution in the initial dataset, this experiment classified the normal category including normal, polyp and low grade with 2031 images and the abnormal category including high grade and cancer with 1974 images. Examples of the HE-CRC-DS are shown in Fig.~\ref{fig:Example-CRC}. The normal category is shown in Fig.~\ref{fig:Example-CRC}-(a). All of them have intact oval glands with neatly arranged nuclei. The abnormal category is shown in Fig.~\ref{fig:Example-CRC}-(b). The boundaries of the glandular structures are not clear and the nuclei are drastically enlarged.
\subsubsection{Data Settings}
All the images in the HE-CRC-DS, including the normal and abnormal categories, are randomly partitioned into training data and test data at a ratio of 1:1. In the training data, the training and validation sets are randomly assigned three times at a ratio of 1:1 and used to perform three randomised experiments. All of these are resized to 224 $\times$ 224 pixels using bilinear interpolation. Because the small size of the medical image dataset leads to a large amount of error information in training, the training set is enlarged to six times by rotating it $90^{\circ}$, $180^{\circ}$, and $270^{\circ}$ and horizontal and vertical mirroring. The biomedical researcher uses the ``MATLAB R2020a" software to perform resizing, rotation, and mirroring operations in the pre-processing stage. The pathologist uses the ``Photoshop" software to label the attention areas in the IL stage. The normal and abnormal categories of the epithelial tissues are labelled to minimise the impact of mesenchymal misclassification. The data settings are listed in Table~\ref{Table:data}.
\begin{table}[ht]
\centering
\caption{Data setting of HE-CRC-DS for training, validation and test sets.}
\label{Table:data}
\begin{tabular}{llll}
\hline
Image Type & Training & Validation & Test \\ \hline
Normal     & 3048     & 508        & 1015 \\
Abnormal   & 2964     & 493        & 987  \\
Sum        & 6012     & 1001       & 2002 \\ \hline
\end{tabular}
\end{table}
\subsubsection{Hyper-parameter Setting}
\label{parametersetting}
The IL-MCAM framework consists of two stages. In the AL stage, MCAM model uses 100 epochs and 16 batch sizes trained by the HE-CRC-DS. In the AL and IL stages, the model parameters preserved in each iteration are those with the highest validation set accuracy in this iteration. It uses a modified transfer learning approach in Section~\ref{subsection:TL} for the CHIC task. In the IL stage, one iteration is set to 50 epochs, and only the last fully connected layer is trained using a fine-tuning approach. The AdamW optimiser~\cite{loshchilov-2018-fixing} is used for optimization, and its parameters are set to $2e-3$ learning rate, $1e-8$ eps, $[0.9,0.999]$ betas and $1e-2$ weight decay.
\subsubsection{Evaluation Criteria}
To overcome the bias between different algorithms, it is crucial to choose the appropriate evaluation criteria. Specificity (Spec.), sensitivity (Sens.), F1-score (F1) and average accuracy (Avg.Acc.) are the most standard metrics for evaluating classification performance. True positive (TP), true negative (TN), false positive (FP) and false negative (FN) are used to define these criteria in Table~\ref{table:criteria}. Spec. represents to the ratio of all negative samples predicted to be correct to all actual negative samples. Sens. represents the ratio of correctly classified positive samples to all actual positive samples. F1 is a comprehensive consideration of precision and recall, and it is a critical evaluation criterion for evaluating a model. Avg.Acc is the most typical and fundamental evaluation criterion.
\begin{table}[ht]
		\small
		\centering
		\renewcommand\arraystretch{1.5}
		\caption{Criteria and corresponding definitions for image classification evaluation.}
	\label{table:criteria}
	\begin{tabular}{llll}%几列就有几个c
		\hline
		Criteria & Definition & Criteria & Definition \\\hline
		Spec. & $\rm \frac{TN}{TN+FP}$ & Sens. & $\rm \frac{TP}{TP+FN}$ \\
		F1 & $\rm \frac{2\times TP}{2\times TP+FP+FN}$ & Avg.Acc. & $\rm \frac{TP+TN}{TP+TN+FP+FP}$\\\hline
	\end{tabular}
\end{table}
\subsection{Classification Evaluation}
\subsubsection{Experimental Results}
\begin{figure*}[htbp]
\centering
\includegraphics[trim={0cm 0cm 0cm 0cm},clip,width= 0.8\textwidth]{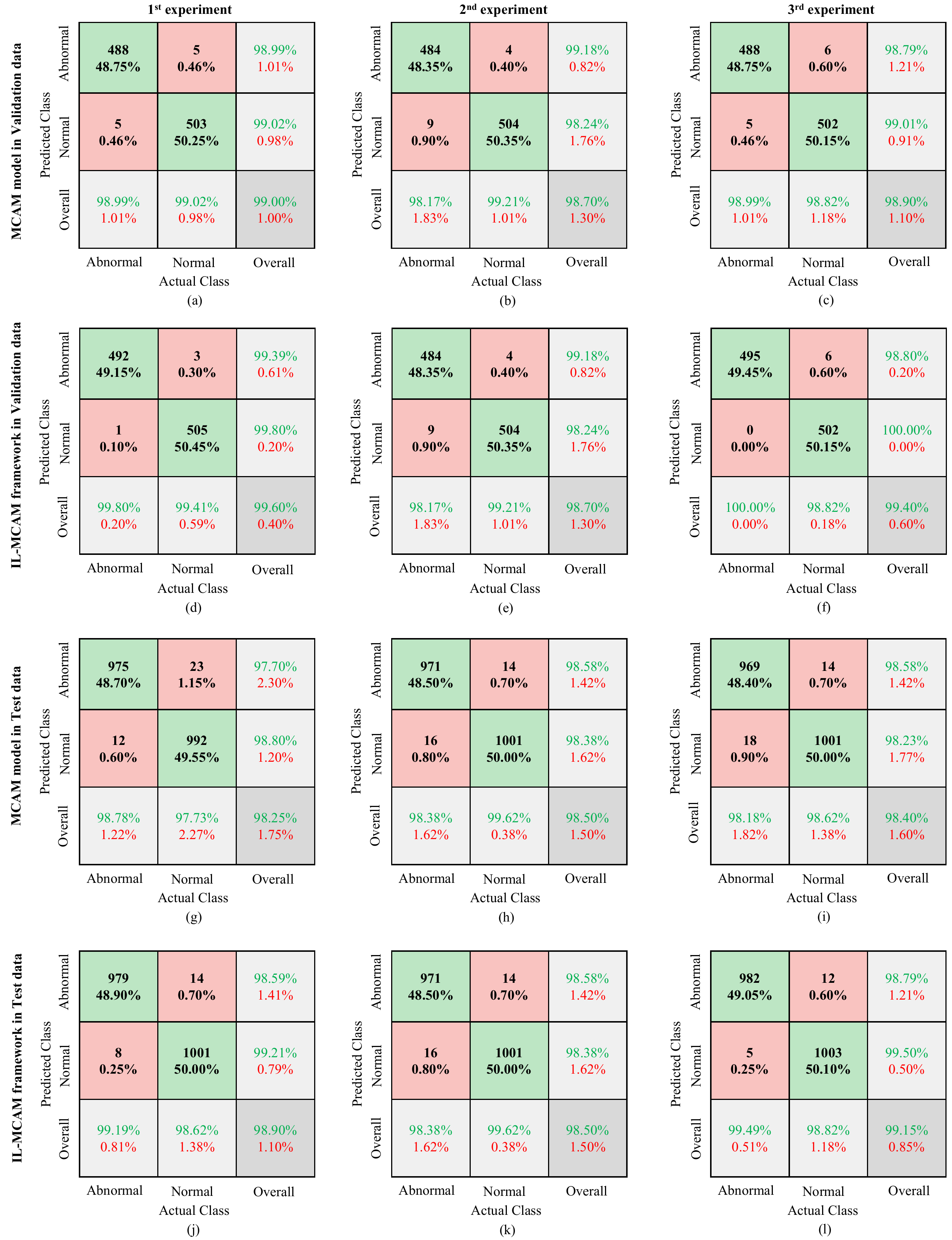}
\caption{Confusion matrix for three randomised experiments in the CHIC task. $1^{st}$ to $4^{th}$ rows are used to represent the results of using MCAM model in validation set, IL-MCAM framework in validation set, MCAM model in test set and IL-MCAM framework in test set, respectively. Each column represents each randomized experiment.}
\label{fig:Confusion}
\end{figure*}

To analyse the experimental results, we show the confusion matrix obtained from three randomised experiments of the proposed MCAM model and IL-MCAM framework in Fig.~\ref{fig:Confusion}. Three randomised experiments and the average evaluation results of the proposed MCAM model and IL-MCAM framework are shown in Table~\ref{Table:result_valtest}.

In the $1^{st}$ experiment, the MCAM model is used in the validation and test sets for classification, and the confusion matrix is shown in Figs.~\ref{fig:Confusion}-(a) and (g). For the validation set, five abnormal category images are misclassified as normal, and five normal category images are misclassified as abnormal. For the test set, 12 abnormal category images are misclassified as normal, and 23 normal category images are misclassified as abnormal. Additionally, the IL-MCAM framework is also used in the validation and test sets for classification, and the confusion matrix is shown in Figs.~\ref{fig:Confusion}-(d) and (j). For the validation set, one abnormal category image is misclassified as normal, and three normal category images are misclassified as abnormal. For the test set, eight abnormal category images are misclassified as normal, and 14 normal category images are misclassified as abnormal. In summary, compared with the MCAM model, the IL-MCAM framework identifies two more correct abnormal images and four more normal images in the validation set and identifies nine more correct abnormal images and four more normal images in the test set.

In the $2^{nd}$ experiment, the MCAM model is used in the validation and test sets for classification, and the confusion matrix is shown in Figs.~\ref{fig:Confusion}-(b) and (h). For the validation set, nine abnormal category images are misclassified as normal, and four normal category images are misclassified as abnormal. For the test set, 16 abnormal category images are misclassified as normal, and 14 normal category images are misclassified as abnormal. Additionally, the IL-MCAM framework is also used in the validation and test sets for classification. The confusion matrix is shown in Figs.~\ref{fig:Confusion}-(e) and (k), and the results are the same as those for the MCAM model. These results suggest that the addition of the IL stage did not occur in this randomised experiment.

In the $3^{rd}$ experiment, the MCAM model is used in the validation and test sets for classification, and the confusion matrix is shown in Figs.~\ref{fig:Confusion}-(c) and (i). For the validation set, five abnormal category images are misclassified as normal, and six normal category images are misclassified as abnormal. For the test set, 18 abnormal category images are misclassified as normal, and 14 normal category images are misclassified as abnormal. Additionally, the IL-MCAM framework is also used in the validation and test sets for classification, and the confusion matrix is shown in Figs.~\ref{fig:Confusion}-(d) and (j). For the validation set, no abnormal category images are misclassified as normal, and six normal category images are misclassified as abnormal. For the test set, five abnormal category images are misclassified as normal, and 12 normal category images are misclassified as abnormal. In summary, compared with the MCAM model, the IL-MCAM framework identifies five more correct abnormal images and no more normal images in the validation set and 13 more correct abnormal images and two more normal images in the test set.

The following results are obtained from Table~\ref{Table:result_valtest}. First, it can be observed that the accuracy of classification results of the first and third randomised experiments using the IL-MCAM framework is higher than that of accuracy using MCAM model, except for the second randomized experiment. In the three randomised experiments, the MCAM model achieves $99.02\%$, $98.72\%$ and $98.85\%$ on average in the abnormal category of the validation set and $98.32\%$, $98.45\%$ and $98.37\%$ on average in the abnormal category of the test set for Spec., Sens. and F1, respectively. Based on the MCAM model, using the IL-MCAM framework improve the Spec., Sens., and F1 by $0.13\%$, $0.60\%$ and $0.37\%$ on average for the abnormal category of the validation set and $0.59\%$, $0.35\%$ and $0.47\%$ for the abnormal category of the test set, respectively. The Avg.Acc. of the three randomised experiments improves by $0.36\%$ and $0.47\%$ in the validation and test sets, respectively. It is observed that using IL-MCAM framework can improve the classification effect of the MCAM model.  Furthermore, in the three randomised experiments, regardless of whether the MCAM model or the IL-MCAM framework is used for classification, the deviations between the accuracies of the validation and test sets are not more than $1.00\%$. This indicates that the IL-MCAM framework has good extensibility and robustness. Finally, the standard deviations of the three randomized experiments obtained using the MCAM model are $0.12\%$ and $0.10\%$ for the validation and test sets, respectively. The standard deviations of the three randomised experiments using the IL-MCAM framework are $0.39\%$ and $0.27\%$ for the validation and test sets, respectively. The slight fluctuation of the standard deviation indicates that the MCAM model and IL-MCAM framework have good stability.

\begin{table*}[htbp]
\small
\centering
\caption{Performance analysis of the proposed MCAM model and IL-MCAM framework on validation and test sets among three randomised experiments. ([In \%].)}
\label{Table:result_valtest}
\begin{tabular}{p{0.17\columnwidth}p{0.17\columnwidth}llllclllc}
\hline
\multirow{2}{*}{Test Name}       & \multirow{2}{*}{\begin{tabular}[c]{@{}l@{}}Model\\ /Framework\end{tabular}} & \multirow{2}{*}{Category} & \multicolumn{4}{l}{Validation Set}              & \multicolumn{4}{l}{Test Set}                  \\ \cline{4-11} 
                                 &                                                                             &                           & Spec.  & Sens.  & F1   & Avg.Acc                & Spec. & Sens. & F1   & Avg.Acc                \\ \hline
\multirow{4}{*}{\begin{tabular}[c]{@{}l@{}}$1^{st}$\\ Experiment\end{tabular}}          & \multirow{2}{*}{MCAM}                                                       & Abnormal                  & 99.02  & 98.99  & 98.99 & \multirow{2}{*}{99.00} & 97.73 & 98.78 & 98.24 & \multirow{2}{*}{98.25} \\
                                 &                                                                             & Normal                    & 98.99  & 99.02  & 99.02 &                        & 98.78 & 97.73 & 98.27 &                        \\
                                 & \multirow{2}{*}{IL-MCAM}                                                    & Abnormal                  & 99.41  & 99.80  & 99.60 & \multirow{2}{*}{99.60} & 98.62 & 99.19 & 98.89 & \multirow{2}{*}{98.90} \\
                                 &                                                                             & Normal                    & 99.80  & 99.41  & 99.60 &                        & 99.19 & 98.62 & 98.91 &                        \\ \hline
\multirow{4}{*}{\begin{tabular}[c]{@{}l@{}}$2^{nd}$\\ Experiment\end{tabular}}          & \multirow{2}{*}{MCAM}                                                       & Abnormal                  & 99.21  & 98.17  & 98.67 & \multirow{2}{*}{98.70} & 98.62 & 98.38 & 98.48 & \multirow{2}{*}{98.50} \\
                                 &                                                                             & Normal                    & 98.17  & 99.21  & 98.73 &                        & 98.38 & 98.62 & 98.52 &                        \\
                                 & \multirow{2}{*}{IL-MCAM}                                                    & Abnormal                  & 99.21  & 98.17  & 98.67 & \multirow{2}{*}{98.70} & 98.62 & 98.38 & 98.48 & \multirow{2}{*}{98.50} \\
                                 &                                                                             & Normal                    & 98.17  & 99.21  & 98.73 &                        & 98.38 & 98.62 & 98.52 &                        \\ \hline
\multirow{4}{*}{\begin{tabular}[c]{@{}l@{}}$3^{rd}$\\ Experiment\end{tabular}}          & \multirow{2}{*}{MCAM}                                                       & Abnormal                  & 98.82  & 98.99  & 98.89 & \multirow{2}{*}{98.90} & 98.62 & 98.18 & 98.38 & \multirow{2}{*}{98.40} \\
                                 &                                                                             & Normal                    & 98.99  & 98.82  & 98.92 &                        & 98.18 & 98.62 & 98.43 &                        \\
                                 & \multirow{2}{*}{IL-MCAM}                                                    & Abnormal                  & 98.82  & 100.00 & 99.39 & \multirow{2}{*}{99.40} & 99.49 & 98.82 & 99.14 & \multirow{2}{*}{99.15} \\
                                 &                                                                             & Normal                    & 100.00 & 98.82  & 99.41 &                        & 98.82 & 99.49 & 99.16 &                        \\ \hline
\multirow{4}{*}{Average} & \multirow{2}{*}{MCAM}                                                       & Abnormal                  & 99.02  & 98.72  & 98.85 & \multirow{2}{*}{$98.87\pm0.12$} & 98.32 & 98.45 & 98.37 & \multirow{2}{*}{$98.38\pm0.10$} \\
                                 &                                                                             & Normal                    & 98.72  & 99.02  & 98.89 &                        & 98.45 & 98.32 & 98.41 &                        \\
                                 & \multirow{2}{*}{IL-MCAM}                                                    & Abnormal                  & 99.15  & 99.32  & 99.22 & \multirow{2}{*}{$99.23\pm0.39$} & 98.91 & 98.80 & 98.84 & \multirow{2}{*}{$98.85\pm0.27$} \\
                                 &                                                                             & Normal                    & 99.32  & 99.15  & 99.25 &                        & 98.80 & 98.91 & 98.86 &                        \\ \hline
\end{tabular}
\end{table*}
\subsubsection{Contrast Experiment of CHIC}

There are three contrast experiments as follows: The first compares the proposed IL-MCAM framework with other traditional deep learning models, the second compares the proposed IL-MCAM framework with models that do not use TL, and the third compares the proposed IL-MCAM framework with models that do not use AM.

\emph{\textbf{Comparison with other deep learning models:}} To validate the excellent performance of the MCAM model and IL-MCAM framework in the CHIC task, we compare 18 different basic deep learning models, including CNN models, vision transformer (VT) models and MLP models, which are AlexNet~\cite{krizhevsky-2012-ICWDC}, VGG-16~\cite{simonyan-2014-VGG}, Inception-V3~\cite{szegedy-2016-inception}, ResNet-50~\cite{he-2016-DRLFI}, Xception~\cite{chollet-2017-xception}, ResNeXt-50~\cite{xie-2017-ARTFD}, InceptionResNet-V1~\cite{szegedy-2017-IIATI}, DenseNet-121~\cite{huang-2017-densenet}, ViT~\cite{dosovitskiy-2020-AIIWW}, DeiT~\cite{touvron-2021-TDITD}, BoTNet-50~\cite{srinivas-2021-BTFVR}, CoaT~\cite{xu-2021-CCIT}, CaiT~\cite{touvron-2021-GDWIT}, T2T-ViT~\cite{yuan-2021-T2TVIT}, LeViT~\cite{graham-2021-levit}, MLP-Mixer~\cite{tolstikhin-2021-MAAAF}, gMLP~\cite{liu-2021-PATM} and ResMLP~\cite{touvron-2021-RFNFI}.

The results of the contrast experiment between the IL-MCAM framework and other deep learning models are shown in Table~\ref{Table:compare}. the evaluation criteria are obtained by averaging the results of three the randomised experiments. The following results are obtained and are listed in Table~\ref{Table:compare} by analysing the performance of traditional deep learning models on the test set.First, Inception-V3, VGG-16 and Inception-V3 have the best Spec., Sens., and F1, respectively, in the abnormal category with values of $98.13\%$, $98.65\%$ and $98.23\%$, respectively. Meanwhile, VGG-16, Inception-V3, and Inception-V3 have the best Spec., Sens. and F1, respectively,  in the normal with values of $98.65\%$, $98.13\%$ and $98.23\%$, respectively. Finally, Inception-V3 has the best Avg.Acc. of $98.25\%$.

Compared with traditional deep learning models, the proposed MCAM model and IL-MCAM framework can improve the performance. Although Sens. obtained by the MCAM model in the abnormal category is $0.20\%$ lower than the optimal Sens. of the traditional deep learning models, it is higher than that of all other deep learning models except the VGG-16. Meanwhile, Spec. and F1 obtained by the MCAM model in the abnormal category are $0.19\%$ and $0.12\%$, respectively, which are better than those of traditional deep learning models. Similarly, the MCAM model improves Spec. and F1 compared to traditional deep learning models in the normal category, except for a slight decrease in Sens. Most importantly, the Avg.Acc. of the MCAM model is $0.13\%$ higher than the optimal result of traditional deep models, indicating that the MCAM model improves the classification performance compared with traditional deep learning models.

Compared with the optimal results obtained by traditional deep learning models, Spec., Sens. and F1 obtained by the IL-MCAM framework in the abnormal category improved by $0.78\%$, $0.15\%$, and $0.61\%$, respectively. In addition, Spec., Sens. and F1 obtained by the IL-MCAM framework in the normal category improved by $0.15\%$, $0.78\%$, and $0.60\%$, respectively, compared with traditional deep learning models. Finally, the Avg.Acc. of the IL-MCAM framework is $0.60\%$ higher than that of traditional deep learning models. The above comparison results indicate that the IL-MCAM framework performs better than traditional deep learning models in the CHIC task.

The contrast experiment result between the IL-MCAM framework and traditional deep learning method shows that the proposed MCAM model has improved significant improvement compared to the traditional deep learning model in the CHIC task. Furthermore, interactive learning of the IL-MCAM framework can further improve the classification performance of the MCAM model.

\begin{table*}[htbp]
\small
\centering
\caption{Performance analysis of the proposed MCAM model and IL-MCAM approach along with the traditional models on test set. ([In \%].)}
\label{Table:compare}
\begin{tabular}{lllllll}
\hline
Type                  & Model/Framework                     & Category & Spec. & Sens. & F1    & Avg.Acc.               \\ \hline
\multirow{16}{*}{CNN} & \multirow{2}{*}{AlexNet~\cite{krizhevsky-2012-ICWDC}}            & Abnormal & 93.50 & 96.35 & 94.90 & \multirow{2}{*}{94.90} \\
                      &                                     & Normal   & 96.35 & 93.50 & 94.90 &                        \\
                      & \multirow{2}{*}{VGG-16~\cite{simonyan-2014-VGG}}             & Abnormal & 96.75 & 98.65 & 97.68 & \multirow{2}{*}{97.68} \\
                      &                                     & Normal   & 98.65 & 96.75 & 97.69 &                        \\
                      & \multirow{2}{*}{Inception-V3~\cite{szegedy-2016-inception}}       & Abnormal & 98.13 & 98.38 & 98.23 & \multirow{2}{*}{98.25} \\
                      &                                     & Normal   & 98.38 & 98.13 & 98.26 &                        \\
                      & \multirow{2}{*}{ResNet-50~\cite{he-2016-DRLFI}}          & Abnormal & 95.76 & 93.21 & 94.36 & \multirow{2}{*}{94.51} \\
                      &                                     & Normal   & 93.21 & 95.76 & 94.65 &                        \\
                      & \multirow{2}{*}{Xception~\cite{chollet-2017-xception}}           & Abnormal & 97.90  & 98.45 & 98.15 & \multirow{2}{*}{98.17} \\
                      &                                     & Normal   & 98.45 & 97.90 & 98.19 &                        \\
                      & \multirow{2}{*}{ResNeXt-50~\cite{xie-2017-ARTFD}}         & Abnormal & 93.40 & 93.82 & 93.54 & \multirow{2}{*}{93.60} \\
                      &                                     & Normal   & 93.82 & 93.40 & 93.68 &                        \\
                      & \multirow{2}{*}{InceptionResNet-V1~\cite{szegedy-2017-IIATI}} & Abnormal & 95.57 & 96.45 & 95.96 & \multirow{2}{*}{96.00} \\
                      &                                     & Normal   & 96.45 & 95.57 & 96.04 &                        \\
                      & \multirow{2}{*}{DenseNet-121~\cite{huang-2017-densenet}}       & Abnormal & 96.16 & 97.67 & 96.90 & \multirow{2}{*}{96.90} \\
                      &                                     & Normal   & 97.67 & 96.16 & 96.90 &                        \\ \hline
\multirow{14}{*}{VT}  & \multirow{2}{*}{ViT~\cite{dosovitskiy-2020-AIIWW}}                & Abnormal & 77.90 & 74.90 & 75.80 & \multirow{2}{*}{76.42} \\
                      &                                     & Normal   & 74.90 & 77.90 & 77.02 &                        \\
                      & \multirow{2}{*}{DeiT~\cite{touvron-2021-TDITD}}               & Abnormal & 94.77 & 92.30 & 93.39 & \multirow{2}{*}{93.56} \\
                      &                                     & Normal   & 92.30 & 94.77 & 93.72 &                        \\
                      & \multirow{2}{*}{BoTNet-50~\cite{srinivas-2021-BTFVR}}          & Abnormal & 94.48 & 95.54 & 94.96 & \multirow{2}{*}{95.01} \\
                      &                                     & Normal   & 95.54 & 94.48 & 95.04 &                        \\
                      & \multirow{2}{*}{CaiT~\cite{touvron-2021-GDWIT}}               & Abnormal & 75.66 & 72.44 & 73.37 & \multirow{2}{*}{74.08} \\
                      &                                     & Normal   & 72.44 & 75.66 & 74.74 &                        \\
                      & \multirow{2}{*}{CoaT~\cite{xu-2021-CCIT}}               & Abnormal & 73.89 & 87.94 & 81.89 & \multirow{2}{*}{80.82} \\
                      &                                     & Normal   & 87.94 & 73.89 & 79.62 &                        \\
                      & \multirow{2}{*}{T2T-ViT~\cite{yuan-2021-T2TVIT}}            & Abnormal & 90.15 & 92.00 & 91.03 & \multirow{2}{*}{91.05} \\
                      &                                     & Normal   & 92.00 & 90.15 & 91.09 &                        \\
                      & \multirow{2}{*}{LeViT~\cite{graham-2021-levit}}              & Abnormal & 79.31 & 82.06 & 80.71 & \multirow{2}{*}{80.66} \\
                      &                                     & Normal   & 82.06 & 79.31 & 80.62 &                        \\ \hline
\multirow{6}{*}{MLP}  & \multirow{2}{*}{MLP-Mixer~\cite{tolstikhin-2021-MAAAF}}          & Abnormal & 73.10 & 72.14 & 72.20 & \multirow{2}{*}{72.63} \\
                      &                                     & Normal   & 72.14 & 73.10 & 73.00 &                        \\
                      & \multirow{2}{*}{gMLP~\cite{liu-2021-PATM}}               & Abnormal & 88.07 & 88.96 & 88.42 & \multirow{2}{*}{88.51} \\
                      &                                     & Normal   & 88.96 & 88.07 & 88.60 &                        \\
                      & \multirow{2}{*}{ResMLP~\cite{touvron-2021-RFNFI}}             & Abnormal & 72.12 & 77.41 & 75.12 & \multirow{2}{*}{74.73} \\
                      &                                     & Normal   & 77.41 & 72.12 & 74.32 &                        \\ \hline
\multirow{4}{*}{Ours} & \multirow{2}{*}{MCAM}               & Abnormal & 98.32 & 98.45 & 98.37 & \multirow{2}{*}{98.38} \\
                      &                                     & Normal   & 98.45 & 98.32 & 98.41 &                        \\
                      & \multirow{2}{*}{IL-MCAM}            & Abnormal & 98.91 & 98.80 & 98.84 & \multirow{2}{*}{98.85} \\
                      &                                     & Normal   & 98.53 & 98.91 & 98.73 &                        \\ \hline
\end{tabular}
\end{table*}

\emph{\textbf{Comparison with IL-MCAM Framework without TL:}} To validate the effectiveness of TL in the experiment, we conduct a comparison experiment between a model using TL and a model without TL during the re-training process, and the experimental results from three randomised experiments are shown in Fig.~ref{fig:compare-freeze}. The model without TL has $98.00\%$, $98.48\%$, and $98.21\%$ of Spec., Sens. and F1, respectively, in the abnormal category. The Spec., Sens. and F1 of the model with TL in the abnormal category are $98.91\%$, $98.80\%$ and $98.84\%$, respectively, which is an improvement of $0.91\%$, $0.32\%$ and $0.63\%$, respectively, compared to the model without TL. The Spec., Sens. and F1 of the model without TL are $98.48\%$, $98.00\%$, and $98.25\%$, respectively, in the normal category. The Spec., Sens. and F1 of the model with TL in the normal category are $98.80\%$, $98.91\%$ and $98.86\%$, respectively, which are $0.32\%$, $0.91\%$ and $0.59\%$ higher than those of the model without TL. The Avg.Acc. for the model without TL is $98.23\%$, whereas the Avg.Acc. of the model using TL is $98.85\%$, which is $0.62\%$ higher than the model without TL. In summary, the comparison experiment illustrates that the proposed IL-MCAM framework using TL is better than the model without TL.

\begin{figure}[ht]
\centering
\includegraphics[trim={0cm 0cm 0cm 0cm},clip,width= 0.48\textwidth]{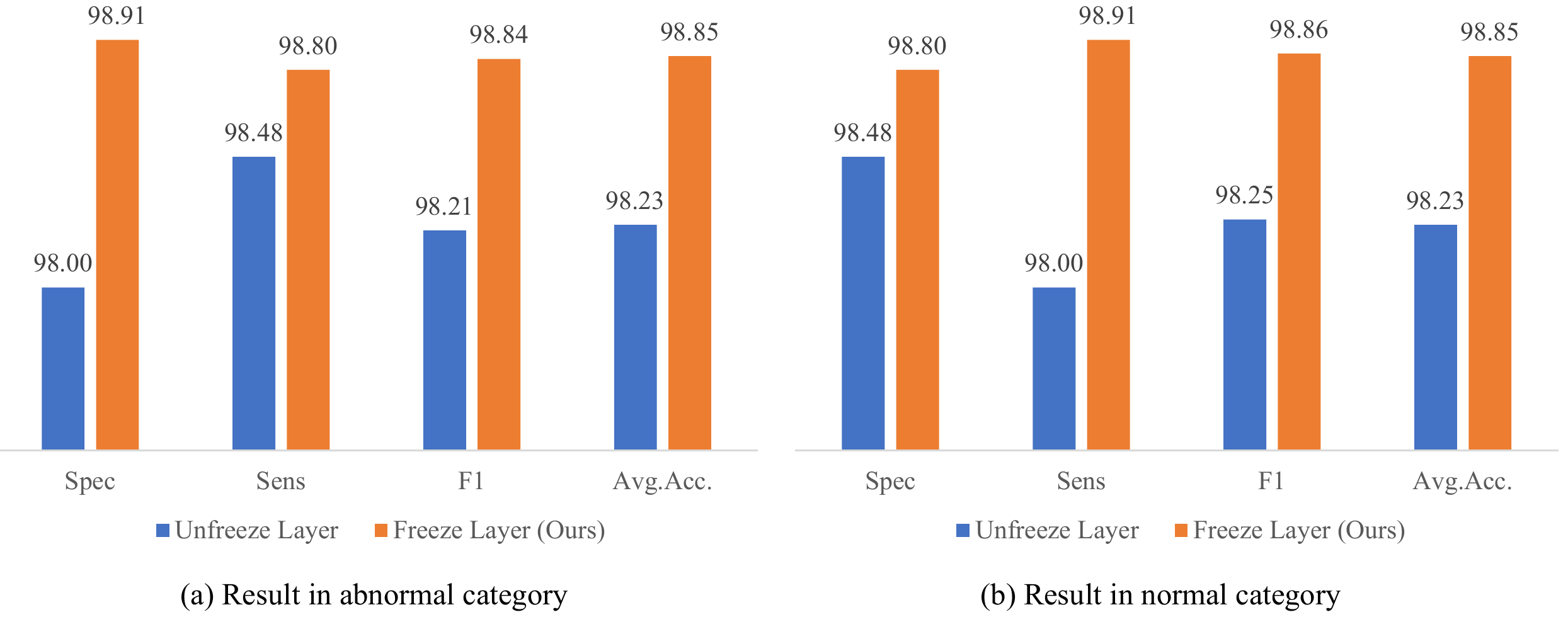}
\caption{Performance analysis about whether to freeze the network layer in AL stage on test set. (a) is the performance in abnormal category. (b) is the performance in normal category. ([In \%].)}
\label{fig:compare-freeze}
\end{figure}

\begin{figure}[ht]
\centering
\includegraphics[trim={0cm 0cm 0cm 0cm},clip,width= 0.48\textwidth]{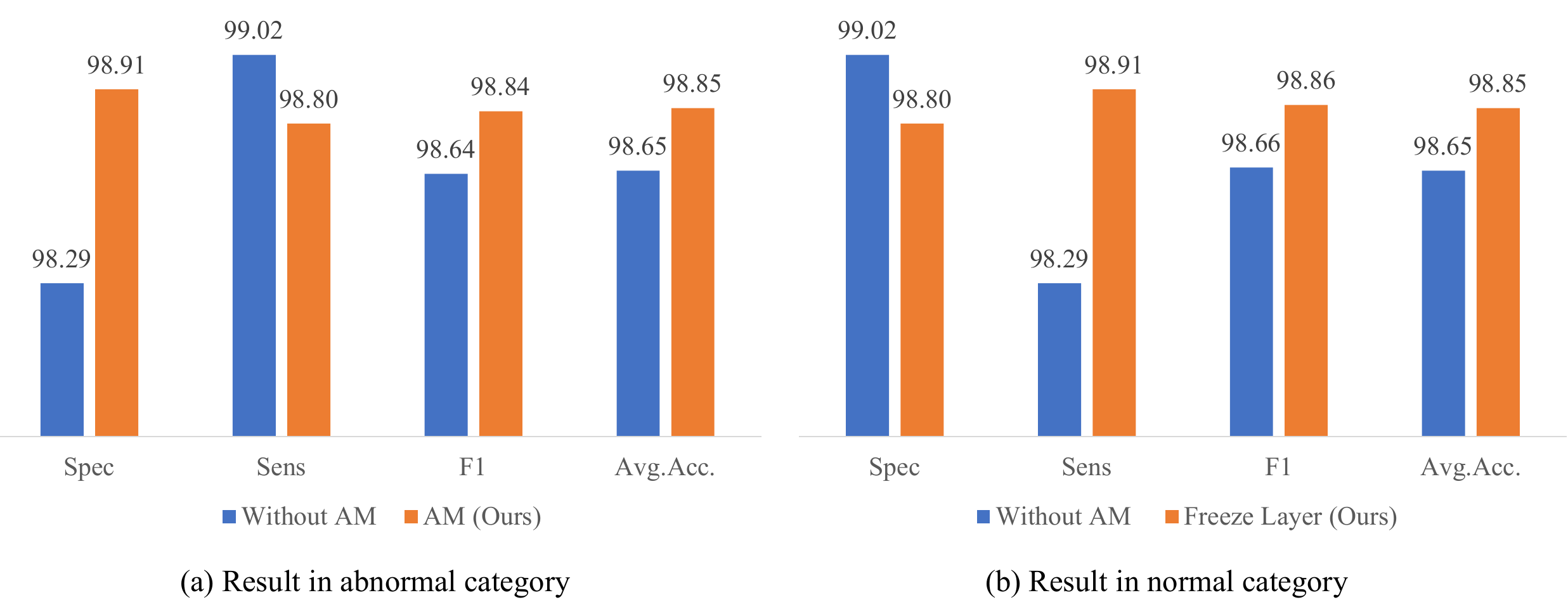}
\caption{Performance analysis about whether to use AM in MCAM model on test set. (a) is the performance in abnormal category. (b) is the performance in normal category. ([In \%].)}
\label{fig:compare-am}
\end{figure}

\emph{\textbf{Comparison with ensemble model without AM:}} To validate the effectiveness of the AM module in the experiment, we replaced SIC, MGIC and MSIC channels with traditional the VGG-16, Inception-V3 and Xception to obtain an ensemble model. The results of the ensemble model and IL-MCAM framework obtains from averages of three randomised experiments, as shown in Fig.~\ref{fig:compare-am}. The ensemble model has $98.29\%$, $99.02\%$ and $98.64\%$ of Spec., Sens. and F1 in the abnormal category. The IL-MCAM framework has $98.91\%$, $98.80\%$ and $98.84\%$ of Spec., Sens. and F1, respectively, in the abnormal category. Although Sens. is lower the IL-MCAM framework than the ensemble model, the most critical evaluation criterion, F1, is $0.20\%$ higher. Similar results are obtained for the normal category. In addition, the Avg.Acc. of IL-MCAM framework is $98.85\%$, which is $0.20\%$ higher than that of ensemble model. In summary, the IL-MCAM framework with added AM modules is more effective than the ensemble model composed of traditional deep learning models.

\subsection{Extended Experiments}
In this section, we describe the three conducted experiments. In Section~\ref{ablation}, we describe the ablation experiments to verify the roles of the SIC, MGIC and MSIC modules in the IL-MCAM framework. In Section~\ref{interchangeability}, we describe the experiment that used deep learning models combined with other AMs to implement the function of SIC, MGIC and MSIC to verify the interchangeability of the IL-MACM framework. In Section~\ref{nct}, we describe the experiment that used the NCT-CRC-HE-100K dataset for multi-classification experiments to verify good generalisation ability of the IL-MACM framework.
\subsubsection{Ablation Experiment}
\label{ablation}

To verify the role of the three channels in the IL-MCAM framework, we conduct ablation experiments according to the experimental setting described in Section~\ref{parametersetting}. We list the results of the three randomised experiments in Table~\ref{Table:ablationexperiments}, and the importance of each channel as follows.

First, through the ablation experiment in the second row, it can be observed that the Avg.Acc. using only MSIC is only $0.10\%$ lower than that of IL-MCAM, and even in the second randomised experiment on the abnormal category Sens. is higher than the results obtained from the IL-MCAM framework. Meanwhile, it can be observed that the Avg.Acc. decreases more by $0.22\%$ when the MSIC is removed from the ablation experiment in the fifth row. Importantly, these results indicate that the MSIC plays an irreplaceable role in the entire IL-MCAM framework.

Second, through the third row of ablation experiments, the Avg.Acc. using only MGIC is $0.46\%$ lower than that of the IL-MCAM framework, and Sens. for the abnormal class in the third randomised experiment is equal to the results obtained from the IL-MCAM framework. Moreover, through the fourth row of the ablation experiment, the same Avg.Acc. is obtained with the MGIC removed as is obtained with the MSIC removed. Importantly, these results indicate that the MGIC plays a crucial role in the entire IL-MCAM framework.

Finally, it can be observed in the ablation experiment in the first row that the Avg.Acc. using only SIC is $0.80\%$ lower than that of the IL-MCAM framework. Through the sixth row of the ablation experiment, there is a slight decrease of $0.08\%$ in the Avg.Acc. after the SIC is removed. There is a slight decrease in the second and third randomised experiments and no decrease in the first randomised experiment, which indicates that the SIC plays role in the overall IL-MCAM framework, but the effect is limited.

\begin{table*}[ht]
\small
\centering
\caption{Results of the ablation experiments on the three channels obtained on test set. (\checkmark indicates that this channel is used. [In \%].)}
\label{Table:ablationexperiments}
\begin{tabular}{p{0.035\columnwidth}p{0.065\columnwidth}p{0.065\columnwidth}lllllllllll}
\hline
\multicolumn{3}{l}{Channel}                                                 & \multirow{2}{*}{Category} & \multicolumn{3}{l}{$1^{st}$ Experiment} & \multicolumn{3}{l}{$2^{nd}$ Experiment} & \multicolumn{3}{l}{$3^{rd}$ Experiment} & \multirow{2}{*}{Avg.acc.} \\ \cline{1-3} \cline{5-13}
SIC                     & MGIC                    & MSIC                    &                           & Sepc.    & Sens.   & F1     & Sepc.    & Sens.   & F1     & Sepc.    & Sens.   & F1     &                           \\ \hline
\multirow{2}{*}{\checkmark} & \multirow{2}{*}{}       & \multirow{2}{*}{}       & Abnormal                  & 97.93   & 98.38  & 98.13  & 97.14   & 97.67  & 97.37  & 98.72   & 98.48  & 98.58  & \multirow{2}{*}{$98.05\pm0.49$}    \\
                        &                         &                         & Normal                    & 98.38   & 97.93  & 98.17  & 97.67   & 97.14  & 97.43  & 98.48   & 98.72  & 98.62  &                           \\ \hline
\multirow{2}{*}{}       & \multirow{2}{*}{\checkmark} & \multirow{2}{*}{}       & Abnormal                  & 98.72   & 99.09  & 98.89  & 98.22   & 98.68  & 98.43  & 99.09   & 98.72  & 98.89  & \multirow{2}{*}{$98.75\pm0.21$}    \\
                        &                         &                         & Normal                    & 99.09   & 98.72  & 98.91  & 98.68   & 98.22  & 98.47  & 98.72   & 99.09  & 98.91  &                           \\ \hline
\multirow{2}{*}{}       & \multirow{2}{*}{}       & \multirow{2}{*}{\checkmark} & Abnormal                  & 97.34   & 98.48  & 98.08  & 98.62   & 97.77  & 98.17  & 98.89   & 98.82  & 98.84  & \multirow{2}{*}{$98.39\pm0.33$}    \\
                        &                         &                         & Normal                    & 98.48   & 97.34  & 98.12  & 97.77   & 98.62  & 98.23  & 98.82   & 98.89  & 98.87  &                           \\ \hline
\multirow{2}{*}{\checkmark} & \multirow{2}{*}{\checkmark} & \multirow{2}{*}{}       & Abnormal                  & 99.21   & 99.80  & 99.50  & 98.48   & 98.23  & 98.33  & 98.82   & 99.09  & 98.94  & \multirow{2}{*}{$98.77\pm0.30$}    \\
                        &                         &                         & Normal                    & 99.80   & 99.21  & 99.50  & 98.23   & 98.48  & 98.37  & 99.09   & 98.82  & 98.96  &                           \\ \hline
\multirow{2}{*}{\checkmark} & \multirow{2}{*}{}       & \multirow{2}{*}{\checkmark} & Abnormal                  & 98.23   & 98.68  & 98.43  & 98.62   & 98.28  & 98.43  & 99.19   & 98.82  & 98.99  & \multirow{2}{*}{$98.63\pm0.25$}    \\
                        &                         &                         & Normal                    & 98.68   & 98.23  & 98.46  & 98.28   & 98.62  & 98.48  & 98.82   & 99.19  & 99.01  &                           \\ \hline
\multirow{2}{*}{}       & \multirow{2}{*}{\checkmark} & \multirow{2}{*}{\checkmark} & Abnormal                  & 99.41   & 99.80   & 99.60  & 98.62   & 98.28  & 98.43  & 99.09   & 98.82  & 98.93  & \multirow{2}{*}{$98.77\pm0.22$}    \\
                        &                         &                         & Normal                    & 99.80   & 99.41  & 99.60  & 98.28   & 98.62  & 98.48  & 98.82   & 99.09  & 98.96  &                           \\ \hline
\multirow{2}{*}{\checkmark} & \multirow{2}{*}{\checkmark} & \multirow{2}{*}{\checkmark} & Abnormal                  & 99.41   & 99.80  & 99.60  & 98.62   & 98.38  & 98.48  & 99.49   & 98.82  & 99.14  & \multirow{2}{*}{$98.85\pm0.26$}    \\
                        &                         &                         & Normal                    & 99.80   & 99.41  & 99.60  & 98.38   & 98.62  & 98.52  & 98.82   & 99.49  & 99.16  &                           \\ \hline
\end{tabular}
\end{table*}
\subsubsection{Interchangeability Experiment}
\label{interchangeability}
To verify that the three modules in the IL-MCAM framework are interchangeable, we conduct the following an extended experiment using the experimental setting described in Section~\ref{parametersetting}.

In SIC, CBAM~\cite{woo2018cbam} is used instead of SimAM~\cite{yang-2021-simam} because CBAM~\cite{woo2018cbam} is similar to SimAM~\cite{yang-2021-simam} in assigning weights to the spatial information of the VGG-16 model. In MGIC, ECA~\cite{wang-2020-eca} and SRM~\cite{lee2019srm} are similar to SE~\cite{hu-2018-SE} and used to assign weights to the channel information to improve the ability of Inception-V3 model to extract multi-scale global information; therefore, ECA~\cite{wang-2020-eca} and SRM~\cite{lee2019srm} are used instead of SE~\cite{hu-2018-SE}. In MSIC, SE~\cite{hu-2018-SE} and SRM~\cite{lee2019srm} are similar to ECA~\cite{wang-2020-eca} and can assign weights to the channel information to improve the ability of Xception model to extract multi-scale global information; therefore, SE~\cite{hu-2018-SE} and SRM~\cite{lee2019srm} are used instead of ECA~\cite{wang-2020-eca}. The results of the extended experiments to verify the interchangeability arelisted in Table~\ref{Table:Interchangeability}. The first to fourth rows are the replaced AM models, and the fifth row is the proposed IL-MCAM framework. We can observe that the classification accuracies of the four replaced models are $98.38\%$ at the highest and $98.08\%$ at the lowest level, which is a variation not more than $0.90\%$ from the IL-MCAM framework and is a tolerable gap. Furthermore, in the second and third randomised experiments, the Spec. and Sens. of some replaced models in the abnormal category are even higher than that of the IL-MCAM framework. In summary, the three channels of the IL-MCAM framework are interchangeable.
\begin{table*}[ht]
\small
\centering
\caption{Performance analysis of interchangeable experiments using different AMs in each of the three channels. ([In \%].)}
\label{Table:Interchangeability}
\begin{tabular}{llllp{0.075\columnwidth}p{0.075\columnwidth}p{0.075\columnwidth}p{0.075\columnwidth}p{0.075\columnwidth}p{0.075\columnwidth}p{0.075\columnwidth}p{0.075\columnwidth}p{0.075\columnwidth}l}
\hline
\multicolumn{3}{l}{Channel}                                          & \multirow{2}{*}{Category} & \multicolumn{3}{l}{$1^{st}$ Experiment} & \multicolumn{3}{l}{$2^{nd}$ Experiment} & \multicolumn{3}{l}{$3^{rd}$ Experiment} & \multirow{2}{*}{Avg.Acc.} \\ \cline{1-3} \cline{5-13}
SIC                    & MGIC                 & MSIC                 &                           & Sepc.    & Sens.   & F1     & Sepc.    & Sens.   & F1     & Sepc.    & Sens.   & F1     &                           \\ \hline
\multirow{2}{*}{CBAM}  & \multirow{2}{*}{ECA} & \multirow{2}{*}{SRM} & Abnormal                  & 96.65   & 98.38  & 97.49  & 98.28   & 98.82  & 98.53  & 98.72   & 99.49  & 99.09  & \multirow{2}{*}{$98.38\pm0.66$}    \\
                       &                      &                      & Normal                    & 98.38   & 96.65  & 97.52  & 98.82   & 98.28  & 98.58  & 99.49   & 98.72  & 99.11  &                           \\ \hline
\multirow{2}{*}{CBAM}  & \multirow{2}{*}{ECA} & \multirow{2}{*}{SE}  & Abnormal                  & 96.94   & 98.89  & 97.89  & 98.81   & 98.38  & 98.58  & 98.72   & 99.29  & 98.99  & \multirow{2}{*}{$98.50\pm0.45$}    \\
                       &                      &                      & Normal                    & 98.89   & 96.94  & 97.91  & 98.38   & 98.81  & 98.62  & 99.29   & 98.72  & 97.52  &                           \\ \hline
\multirow{2}{*}{CBAM}  & \multirow{2}{*}{SRM} & \multirow{2}{*}{SRM} & Abnormal                  & 97.04   & 98.28  & 97.64  & 98.92   & 98.08  & 98.47  & 97.93   & 99.39  & 98.64  & \multirow{2}{*}{$98.27\pm0.44$}    \\
                       &                      &                      & Normal                    & 98.28   & 97.04  & 97.67  & 98.08   & 98.92  & 98.53  & 99.39   & 97.93  & 98.66  &                           \\ \hline
\multirow{2}{*}{CBAM}  & \multirow{2}{*}{SRM} & \multirow{2}{*}{SE}  & Abnormal                  & 96.55   & 98.78  & 97.65  & 97.64   & 98.18  & 97.88  & 98.62   & 98.78  & 98.68  & \multirow{2}{*}{$98.08\pm0.44$}    \\
                       &                      &                      & Normal                    & 98.78   & 96.55  & 97.66  & 98.18   & 97.64  & 97.93  & 98.78   & 98.62  & 98.72  &                           \\ \hline
\multirow{2}{*}{SimAM} & \multirow{2}{*}{SE}  & \multirow{2}{*}{ECA} & Abnormal                  & 98.62   & 99.19  & 98.89  & 98.62   & 98.38  & 98.48  & 99.49   & 98.82  & 99.14  & \multirow{2}{*}{$98.85\pm0.26$}    \\
                       &                      &                      & Normal                    & 99.19   & 98.62  & 98.91  & 98.38   & 98.62  & 98.52  & 98.82   & 99.49  & 99.16  &                           \\ \hline
\end{tabular}
\end{table*}
\subsubsection{NCT-CRC-HE-100K Image Classification}
\label{nct}
\begin{figure}[ht]
\centering
\includegraphics[trim={0cm 0cm 0cm 0cm},clip,width= 0.45\textwidth]{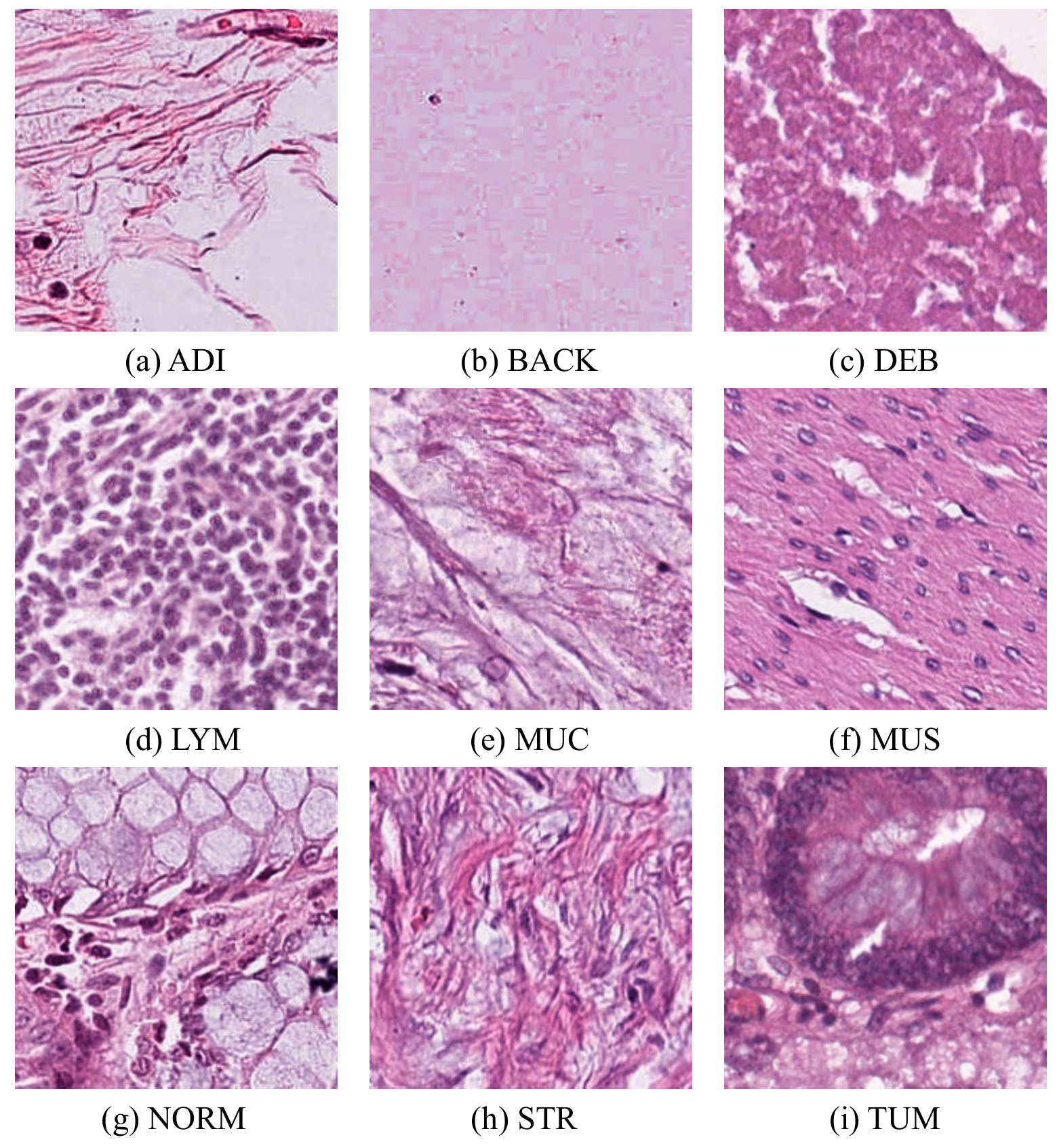}
\caption{Some example of NCT-CRC-HE-100K.}     
\label{fig:example-nct}
\end{figure}

To verify that the IL-MCAM framework has good generalisation ability, we carry out experiments on the publicly available CRC dataset NCT-CRC-HE-100K, composed of 100,000 patch-level images of nine different tissue categories, all of which are 224$\times$224 pixels containing 0.5 microns per pixel. All images are colour-normalized using Macenko's method~\cite{macenko-2009-macenko,knikolas2018crc}. The nine different categories of tissue are adipose (ADI), background (BACK), debris (DEB), lymphocytes (LYM), mucus (MUC), smooth muscle (MUS), normal colon mucosa (NORM), cancer-associated stroma (STR) and colorectal adenocarcinoma epithelium (TUM), some of which are shown in Fig.~\ref{fig:example-nct}. We divide the data in the ratio of 6:2:2 for the training, validation and test sets as shown in Table ~\ref{Table:data-nct}, and conduct extended experiments using the experimental parameters listed in Section~\ref{parametersetting}.

\begin{table}[ht]
\centering
\caption{Data setting of HE-CRC-DS for training, validation and test sets.}
\label{Table:data-nct}
\begin{tabular}{llll}
\hline
Image Type & Training & Validation & Test  \\ \hline
ADI        & 6245     & 2081       & 2081  \\
BACK       & 6340     & 2113       & 2113  \\
DEB        & 6908     & 2302       & 2302  \\
LYM        & 6935     & 2311       & 2311  \\
MUC        & 5338     & 1779       & 1779  \\
MUS        & 8122     & 2707       & 2707  \\
NORM       & 5258     & 1753       & 1752  \\
STR        & 6268     & 2089       & 2089  \\
TUM        & 8591     & 2863       & 2863  \\
Sum        & 60005    & 19998      & 19997 \\ \hline
\end{tabular}
\end{table}

\begin{figure*}[ht]
\centering
\includegraphics[trim={0cm 0cm 0cm 0cm},clip,width= 0.98\textwidth]{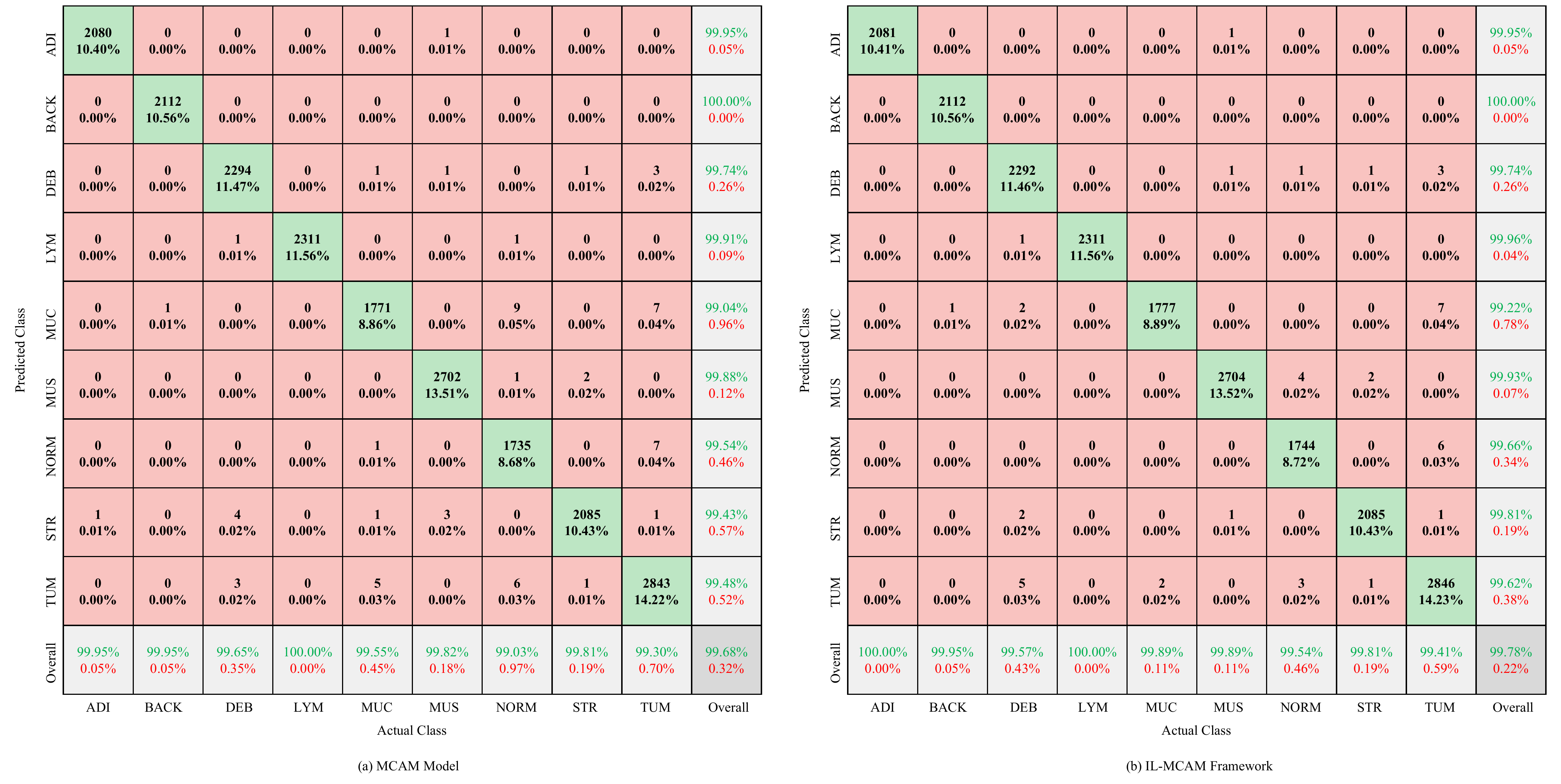}
\caption{Confusion matrix obtained using MCAM model and IL-MCAM framework on test set of NCT-CRC-100K dataset.}     
\label{fig:confusion-nct}
\end{figure*}

The confusion matrix obtained using the MCAM model and the IL-MCAM framework on the test set of the NCT-CRC-100K dataset is shown in Fig.~\ref{fig:confusion-nct}. When using the MCAM model for classification, 19933 images are correctly classified, 64 images are incorrectly classified, and $99.68\%$ of Avg.Acc. is obtained. When using the IL-MCAM model for classification, 19952 images are correctly classified, 45 images are incorrectly classified, and $99.78\%$ of Avg.Acc. is obtained. Compared with the classification results obtained by the MCAM model, IL-MCAM framework improves the classification accuracy in every category except for two fewer images correctly classified in the DEB category.

Finally, we compare our method with previous experimental results obtained in NCT-CRC-HE-100K, and the comparison results are shown in Table~\ref{Table:nct-compare}. The best results in recent years are obtained by the method proposed by Ghosh et al. using the TL and ResNet methods, which obtained $99.76\%$ Avg.Acc. Our proposed IL-MCAM framework obtained $99.78\%$ Avg.Acc. which is $0.02\%$ higher than the previous best result. This result indicates that the proposed IL-MCAM framework exhibits good classification performance and generalisation ability.

\begin{table}[ht]
\centering
\caption{Comparison of the average accuracy of the proposed method with other methods in NCT-CRC-HE-100K. ([In \%].)}
\label{Table:nct-compare}
\begin{tabular}{lll}
\hline
References                             & Methods         & Avg.Acc.       \\ \hline
Kather et al. \cite{kather2019predicting} & TL+VGG-16          & 98.70          \\
Ghosh et al. \cite{ghosh2021colorectal}   & Ensemble CNN          & 96.16          \\
Hamida et al. \cite{hamida2021deep}  & TL+ResNet          & 99.76          \\
\textbf{Proposed}                      & \textbf{MCAM} & \textbf{99.68} \\
\textbf{Proposed}                      & \textbf{IL-MCAM} & \textbf{99.78} \\ \hline
\end{tabular}
\end{table}

\subsection{Experimental Environment and Computational Time}
In our experiments, the proposed IL-MCAM framework have an AL stage and an IL stage. In the AL stage, it took 1.23h to train the MCAM model in parallel. In IL stage, it took 5 min to label each misclassified image and 40 min to fine-tune the model. This experiment is carried out on a workstation. The running memory of the workstation is 32GB. It uses the Windows 10 Professional operating system and is equipped with an 8GB NVIDIA GeForce RTX 4000 GPU. Python 3.6, Pytorch 1.7.0, and Torchvision 0.9.0 are configured on the workstation.

\section{Discussion}\label{section:ds}
This year, the rapid development of deep learning models has played a crucial role in the field of medical diagnosis. Classification of colorectal histopathological images plays a crucial role in the early prevention of diseases. In this paper, the proposed IL-MCAM framework is used for the classification of HE-CRC-DS and achieves good results.

Compared with regular images, medical images tend to be larger in size and the distribution of focused attention regions of the same class in medical images is not uniform in shape. Traditional CNN models using convolutional kernels tend to overconcentrate computational power on extracting edge information; therefore, we consider using a multi-channel approach combined with an attention mechanism to extract multi-scale information. VGG-16, Inception-V3 and Xception models are generally considered to have a good ability to extract spatial information, multi-scale global information and multi-scale local information, and the combination of SimAM, SE and ECA attention mechanisms further improves the recognition accuracy. The IL-MCAM framework uses three channels, SIC, MSIC and MGIC, to enhance the width and ensure the complementarity of the extracted information. Meanwhile, three AMs are used to enhance the depth of the model to ensure the accuracy of the extracted information in each channel. The IL-MCAM framework enhances the classification performance in terms of width and depth. In summary, we select the models mentioned above to form the MCAM model. 

\begin{figure*}[bp]
\centering
\includegraphics[trim={0cm 0cm 0cm 0cm},clip,width= \textwidth]{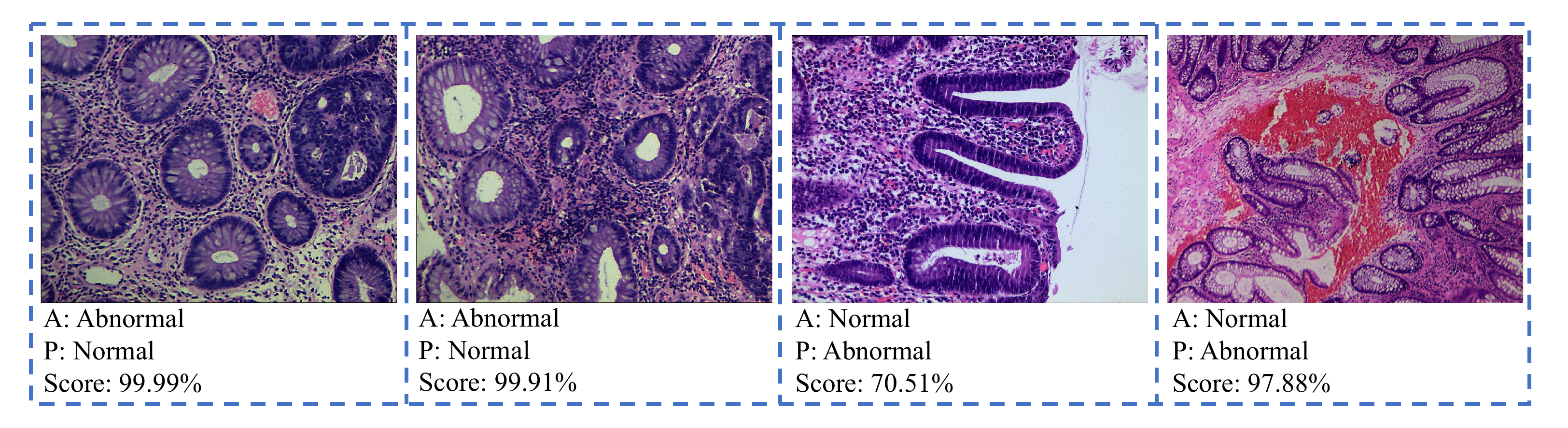}
\caption{Examples of misclassified images from HE-CRC-DS.}
\label{fig:misclass}
\end{figure*}

Table~\ref{Table:size} shows the model parameters and training time for comparing the proposed approaches with other traditional deep learning models. First, we can observe that the proposed MCAM model has very good results and has a significant improvement in classification results compared to traditional automatic methods using interactions. In addition, although the VT and MLP models are more effective than CNN models for routine tasks and have been shown to have a good ability to extract global information, these models do not work well in this experiment because of overfitting. The small medical training set leads to overfitting when trained on a complex or large model, and the experimental results validate this conclusion. In VT models, ViT and CaiT have large model parameters, but the experimental results are not satisfactory, and there are good classification results obtained by the lightweight DeiT and T2T-ViT. The same results are also obtained by the MLP models. Finally, owing to the complexity of the computational process caused by the complexity of the network, some small-scale models also require considerable computation time. In contrast, three channels of SIC, MGIC and MSIC only use simple convolutional and AM blocks, and using parallel training techniques can significantly reduce computation the time of the three channels, so the IL-MCAM framework does not consume a lot of time for training although there are large model parameters.

\begin{table}[ht]
\centering
\caption{Model parameters and training time of comparing between the proposed approaches and other traditional deep learning models.}
\label{Table:size}
\begin{tabular}{lll}
\hline
\multirow{2}{*}{\begin{tabular}[c]{@{}l@{}}Model\\ /Framework\end{tabular}} & \multirow{2}{*}{Size (MB)} & \multirow{2}{*}{Time (s)} \\
                                                                            &                            &                           \\ \hline
AlexNet~\cite{krizhevsky-2012-ICWDC}                                                                     & 217                        & 1331                      \\
VGG-16~\cite{simonyan-2014-VGG}                                                                      & 512                        & 7060                      \\
Inception-V3~\cite{szegedy-2016-inception}                                                                & 83.4                       & 5340                      \\
ResNet-50~\cite{he-2016-DRLFI}                                                                   & 90                         & 4772                      \\
Xception~\cite{chollet-2017-xception}                                                                    & 79.6                       & 4015                      \\
ResNeXt-50~\cite{xie-2017-ARTFD}                                                                  & 88                         & 4564                      \\
InceptionResNet-V1~\cite{szegedy-2017-IIATI}                                                          & 30.8                       & 3260                      \\
DenseNet-121~\cite{huang-2017-densenet}                                                                & 27.1                       & 2860                      \\
ViT~\cite{dosovitskiy-2020-AIIWW}                                                                         & 31.2                       & 1502                      \\
DeiT~\cite{touvron-2021-TDITD}                                                                        & 21.1                       & 2566                      \\
BoTNet-50~\cite{srinivas-2021-BTFVR}                                                                   & 72.1                       & 4772                      \\
CaiT~\cite{touvron-2021-GDWIT}                                                                        & 460                        & 6956                      \\
CoaT~\cite{xu-2021-CCIT}                                                                        & 20.6                       & 3073                      \\
T2T-ViT~\cite{yuan-2021-T2TVIT}                                                                     & 15.5                       & 2852                      \\
LeViT~\cite{graham-2021-levit}                                                                       & 65.8                       & 2943                      \\
MLP-Mixer~\cite{tolstikhin-2021-MAAAF}                                                                   & 225                        & 11284                     \\
gMLP~\cite{liu-2021-PATM}                                                                        & 73.2                       & 6396                      \\
ResMLP~\cite{touvron-2021-RFNFI}                                                                      & 169                        & 8943                      \\
MCAM                                                                        & 639                        & 7060                      \\
IL-MCAM                                                                     & 639                        & 7060                      \\ \hline
\end{tabular}
\end{table}

The confusion matrix for the three randomised experiments is shown in Fig.~\ref{fig:Confusion}. To further analyse the causes of misclassification, we consulted the pathologists in detail and concluded the following. Examples of misclassified images from the three randomised experiments are shown in Fig.~\ref{fig:misclass}. These examples can explain the three main reasons for the misclassification of HE-CRC-DS in the CHIC task using the proposed IL-MCAM framework. First, most of the lumen structure in Figs.~\ref{fig:misclass}-(a) and (b) is regular, and the cancer part occupies a small portion at the edge of the image, so the IL-MCAM framework classifies this image as normal during the testing phase. Second, Fig.~\ref{fig:misclass}-(c) is an image in the low grade category, where the nuclei of some of the luminal structures have started to enlarge, leading the IL-MCAM framework to classify the image as abnormal. Finally, Fig.~\ref{fig:misclass}-(d) shows an image from the normal category with blebs, which is misclassified owing to the presence of blebs.

\section{Conclusion and Future Work}
\label{section:c}
In this paper, we propose an IL-MCAM framework based on attention mechanisms and interactive learning for CHICs. The proposed IL-MCAM framework uses an MCAM model that combined different attention mechanisms for automatic learning. After automatic learning, the misclassified images are iteratively trained by manually labelling the attention regions to achieve the interactive process. Finally, evaluation metrics are obtained by testing. In the CHIC task, a significant performance improvement is observed in the proposed IL-MCAM approach compared with traditional deep learning models. In addition, we conduct three extended experiments: ablation experiments illustrate the role of each channel in the IL-MCAM framework; interchangeability experiments demonstrate the feasibility of designing three channels, and illustrate the interchangeability of the three channels, and extended experiments on the NCT-CRC-HE-100K dataset illustrate the generalisation ability of the IL-MCAM framework.

In the future, to accommodate different CHIC tasks, we plan to find the most suitable model for the current task from attention mechanisms and deep learning models using permutation and combination. We also plan to add attention mechanisms at different locations of deep learning models to analyse the impact of convolutional layers on classification performance in CHIC tasks.

\section*{Declaration of competing interest} 
The authors declare that they have no conflict of interest.

\section*{Acknowledgments}

This work is supported by the ``National Natural Science Foundation of China'' (No.61806047) 
and the ``Fundamental Research Funds for the Central Universities'' (No. N2019003). 
We thank Miss Zixian Li and Mr. Guoxian Li for their important discussion.

\bibliography{mybibfile}

\end{document}